%% file: main.tex
\documentclass[3p,preprint,12pt]{elsarticle}

\usepackage{lineno,hyperref}
\usepackage{amssymb}
\usepackage{adjustbox}
\usepackage{subcaption}
\usepackage{xspace}
\usepackage{xcolor}
\usepackage{soul}
\usepackage{lipsum}
\usepackage[noend]{algpseudocode}
\usepackage{pdfpages}
\usepackage{makecell}
\usepackage{array}
\usepackage[linesnumbered,ruled,vlined,noend]{algorithm2e}
\SetKwInput{KwInput}{Input}  
\SetKwInput{KwOutput}{Output}
\SetKw{KwBy}{by}
\SetKw{Or}{\hspace{\algoskipindent}\itshape or\;}
\SetKwBlock{Condition}{}{}
\newcommand{\subsubsubsection}[1]{\paragraph{#1}\mbox{}\\}
\journal{Journal of Information Sciences Templates}

\newcommand{\MethodName}{MARCO-GE\xspace}
\begin{document}
% \linenumbers
% \linenumbers
\begin{frontmatter}

\title{Automatic selection of clustering algorithms using supervised graph embedding}
% \tnotetext[mytitlenote]{Fully documented templates are available in the elsarticle package on \href{http://www.ctan.org/tex-archive/macros/latex/contrib/elsarticle}{CTAN}.}

\author{Noy Cohen-Shapira \corref{mycorrespondingauthor}}
\cortext[mycorrespondingauthor]{Corresponding author}
\address{Ben-Gurion University of the Negev, Beer-Sheva, Israel}

\ead{noycohe@post.bgu.ac.il} 
\author{Lior Rokach}
\address{Ben-Gurion University of the Negev, Beer-Sheva, Israel}
\ead{liorrk@post.bgu.ac.il} 

% %% or include affiliations in footnotes:
% \author[ben go,mysecondaryaddress]{Elsevier Inc}

% \author[mysecondaryaddress]{Global Customer Service\corref{mycorrespondingauthor}}
% \cortext[mycorrespondingauthor]{Corresponding author}

% \address[mymainaddress]{1600 John F Kennedy Boulevard, Philadelphia}
% \address[mysecondaryaddress]{360 Park Avenue South, New York}

\begin{abstract}

The widespread adoption of machine learning (ML) techniques and the extensive expertise required to apply them have led to increased interest in automated ML solutions that  reduce the need for human intervention. One of the main challenges in applying ML to previously unseen problems is algorithm selection - the identification of high-performing algorithm(s) for a given dataset, task, and evaluation measure. This study addresses the algorithm selection challenge for data clustering,  a fundamental task in data mining that is aimed at grouping similar objects. We present \MethodName, a novel meta-learning approach for the automated recommendation of clustering algorithms. \MethodName first transforms datasets into graphs and then utilizes a graph convolutional neural network technique to extract their latent representation. Using the embedding representations obtained, \MethodName trains a ranking meta-model capable of accurately recommending top-performing algorithms for a new dataset and clustering evaluation measure. An extensive evaluation on 210 datasets, {17} clustering algorithms, and 10 clustering measures demonstrates the effectiveness of our approach and its superiority in terms of predictive and generalization performance over state-of-the-art clustering meta-learning approaches. 

\end{abstract}


\end{frontmatter}

% \linenumbers
\input{Introduction}

\input{RelatedWork}
\input{Method/Method_main}

\input{Experiments/Experiments_average}

\input{Experiments/Experiments_individual_index}

\input{Experiments/Popularity_results}

\input{Experiments/PCA}

\input{Experiments/Parameter_sensitivity}

\newpage
\input{Conclusions}

\bibliography{ref}

\end{document}

%% file: Introduction.tex
\section{Introduction}
% They typically represent users
% and items in a low-dimensional embedding space and then feed the
% embeddings into the following deep network structures for prediction
% Clustering is an important unsupervised data mining task that aim to group similar objects together.
% Data clustering is an important data mining technology that plays a crucial role in numerous scientific applications

Clustering, in which a set of objects is divided into groups of similar objects \cite{jain2010data}, is an important unsupervised learning task used in many fields, including image analysis, document categorization, bioinformatics, and customer segmentation \cite{hruschka2009survey, zakrzewska2005clustering}.
% such that objects within the same cluster are similar
% while objects in different clusters are distinct.
Although a large variety of clustering algorithms have been proposed, there are no guidelines or recommendations for the selection of an appropriate algorithm for a given dataset and target evaluation measure. Thus, human expertise is often required to provide insights about the properties of a specific domain and the characteristics of the various algorithms and their configurations.

Algorithm selection is the task of identifying algorithms that are likely to produce the best performance on a given combination of dataset, task, and evaluation metrics \cite{feurer2015efficient}. A key challenge in applying machine learning (ML) to a previously unseen dataset
is algorithm selection.

The difficulty in the algorithm selection task stems from the inherent characteristics of the dataset, including its size, the number of features and their composition, etc., which affect an algorithm's performance. The high computational cost of testing multiple configurations that include a large set of ML algorithms and their corresponding hyperparameters has driven the need to automate this process.

In recent years, various approaches have been proposed to address  automated machine learning (AutoML) \cite{thornton2013auto, drori2018alphad3m}. The term AutoML is usually used to describe systems  aimed at automating different aspects of the ML process. Meta-learning is an AutoML approach used to deal with the algorithm selection process. 
Meta-learning methods learn the relationship between the learning algorithm and dataset features, in order to identify the features that contribute to the improved performance of one algorithm over another \cite{das2016meta}. While prior studies proposed meta-learning techniques to address algorithm selection for supervised tasks, such as classification \cite{cohen2019autogrd, brazdil2003ranking}, few studies have focused on unsupervised learning problems, and more specifically, on clustering problems \cite{de2008ranking, ferrari2015clustering, pimentel2019new}. We argue that although existing state-of-the-art approaches based on predefined features are simple, they are not necessarily optimized for algorithm selection. In contrast, we use a supervised graph embedding method, which aims to identify the optimal representation for clustering algorithm selection.

We present \MethodName (Meta-learning Approach for Recommending Clustering algorithms by Graph Embedding),  a novel   meta-learning approach based on supervised graph embedding which is optimized for clustering algorithm selection.  \MethodName first converts the interactions of the  dataset's instances into a graph. Next, a graph convolutional neural network technique is utilized in order to produce an embedding representation of the graph. This representation is then used to train a ranking meta-model capable of recommending high-performing algorithms for a previously unseen dataset and clustering evaluation measure. The code of the \MethodName algorithm is publicly available\footnote{https://github.com/noycohen100/MARCO-GE}.

To validate our method, we conduct a comprehensive evaluation on 210 datasets, {17} clustering algorithms, and 10 clustering evaluation measures. All of the measures are internal indices, namely, they assess the performance of the clustering algorithms without a priori knowledge about the clustering problem solution.
Our experiments show that \MethodName outperforms other state-of-the-art meta-learning methods on the task of clustering algorithm selection. 

Our contributions in this study are as follows: 
\begin{itemize}
   \item We introduce an efficient and highly accurate meta-learning approach for the automated selection of clustering algorithms, using a graph convolutional neural network technique optimized for clustering algorithm selection as a tool for dataset representation.     % \item  We introduce graph convolutional neural networks as a tool to analyze datasets and present an efficient and highly accurate meta-learning approach for automatic clustering algorithm selection for any data type.
    
    \item We empirically demonstrate the merits of our approach on a large set of datasets that were analyzed by multiple clustering algorithms and clustering measures. Our results show that \MethodName significantly outperforms existing state-of-the-art approaches on the task of clustering algorithm selection.
    
    \item We demonstrate our method's ability to consistently recommend high-performing algorithms, produce a high-quality recommendation list, and generate a more robust recommendation model; we also show that our method outperforms other state-of-the-art methods.
\end{itemize}

 The rest of the paper is arranged as follows. Section 2 presents an overview of the algorithm selection problem and several meta-learning approaches for clustering algorithm selection. In section 3, a simple popularity-based baseline is proposed, and \MethodName's methodology is described. We present our evaluation in section 4 and discuss the experimental results in section 5.  In section 6, we conclude and highlight some directions for future work.

%% file: RelatedWork.tex
\section{Related work}
\subsection{Algorithm selection using meta-learning}
Selecting an algorithm that is likely to perform well on a given combination of dataset, task, and evaluation measure is one of the main challenges in ML.
The problem of algorithm selection can be formulated as follows: given a set {\textbf{$A$}= $\{a_1,...a_n\}$} of learning algorithms, a  dataset $d\in D$ {(where $D$ is a set of datasets)}, a task $T$ (e.g., classification, clustering, etc.), and an evaluation measure {$m$}, the goal is to find an algorithm {$a^*\in A$} that minimizes or maximizes {$m$} \cite{Muoz2015AlgorithmSF}. 
% Rice \cite{rice1976algorithm} who defined the algorithm selection problem, %proposes
% claimed
% that there is a relation between a learning algorithm performance characteristics. 

% The growing number of ML algorithms, as well as  their hyperparameters, produces exponenet number of configurations combinations , that %make the algorithm selection task even more challenging since 
% turn the algorithm selection process into  a challenge that is infeasible to address with %infeasible task 
% brute force search.% becomes infeasible.
The growing number of ML algorithms, as well as  their hyperparameters, produces an exponential number of configuration combinations, which turns the algorithm selection process into a challenge that is difficult to address with 
brute-force search.
  Meta-learning is one of the main approaches for dealing with the time-consuming nature and high computational cost of testing multiple configurations. By learning about the behavior of ML algorithms and which attributes of a dataset contribute to the improved performance of one algorithm over others (i.e., meta-knowledge), meta-learning is capable of identifying high-performing algorithms for previously unseen datasets \cite{das2016meta}.

Considering the algorithm selection problem, each meta-example in the meta-knowledge represents an ML task that consists of: (1) dataset characteristics, called meta-features; and (2) information about the performance of the algorithm(s) when applied to the dataset. Based on the meta-knowledge, a learning algorithm (i.e., a meta-learner) is trained and generates a meta-model. Then, given the meta-features of a new dataset, the meta-model is capable of recommending  algorithm(s) for that dataset.

Since the dataset's characterization is crucial for effectively learning the relationship between a dataset and an algorithm's performance, it is important to produce significant and meaningful meta-features. These meta-features can then be used to better identify similarities and differences among datasets.
The meta-features proposed in prior research are commonly used for classification problems, and they are divided into three main categories \cite{alcobacca2020mfe}: 
\newline
\textbf{1) Statistical and information-theoretic meta-features}  are derived directly from the dataset. These meta-features can describe simple information (e.g., number of instances in a dataset), statistical measures (e.g., mean, standard deviation), or information theory metrics (e.g., entropy).
\newline
\textbf{2) Model-based meta-features} describe the learning model to be applied on a given dataset, e.g., the number of leaf nodes or the depth of a decision tree.
% are characteristics induced by applying a learning model to a dataset; for example, a given decision tree model can be used to learn about the decision tree model's structural properties, e.g.,  the depth of the tree or the number of leaf nodes.  
\newline
\textbf{3) Landmarking meta-features} are generated by using the estimated performance of a simple learning algorithm on a given dataset for a quick performance assessment.

% To obtain the second component of each meta-example in the meta-knowledge, namely the algorithms' performance, two main categories existing for assessing the clustering algorithms performances: external measures and internal measures. The fundamental difference is whether the evaluation used external knowledge or not. 

The creation of the meta-knowledge is followed by the training process discussed below.
In most cases, the training process can be classified into two main tasks: classification and recommendation. While the classification task is concerned with learning the association of a single class value (i.e., one of the ML algorithms) with each instance, the recommendation task aims at generating a ranking of all of the possible ML candidates for each instance. In this work, we focus on the second task (i.e., recommendation). In the training process, a meta-learner is applied to the training set, i.e., to the meta-knowledge, and produces a meta-model.  In the case of the recommendation task, given the meta-model and a new dataset, the meta-model generates a ranked list of algorithms, ordered by their predicted performance.  
% Usually, an adapted version of the k-Nearest Neighbors (k-NN) algorithm is adopted as a meta-learner \cite{brazdil2008metalearning}. Given the meta-features of a previously unseen dataset, the adapted k-NN selects the k-instances that are most similar to the new dataset. Then, the k corresponding rankings are aggregated.
% % by the Average Ranking (AR) method. 
% % The AR measure can be defined as follows:
% % \begin{equation}
% %     \overline{r}_j = \frac{1}{k}\sum_{i=1}^{k}{r_{ij}} 
% %     \label{average_ranking}
% % \end{equation}
% % where $r_{ij}$ is the rank position of the $j$-th algorithm on dataset $i$, k is the number of most similar datasets, and $\overline{r}_j$ is the average rank position value for the $j$-th algorithm. 

% Finally, the algorithms are ordered by their ranking, such that the first and the last positions are given to the lowest $\overline{r}_j$ value and to the highest value, respectively.

\subsection{Clustering algorithm selection using meta-learning}
While the majority of meta-learning techniques focus on selecting the best algorithm(s) for classification and regression tasks \cite{alcobacca2020mfe, cohen2019autogrd}, few studies are found in the literature for the clustering task.
In the subsections that follow, we briefly present the techniques for evaluating the performance of clustering algorithms and discuss four studies addressing the clustering algorithm selection problem.

\subsubsection{Clustering algorithm performance assessment} \label{clustering_performance_assessment}

There are two main ways to assess the performance of clustering algorithms: external measures and internal measures. The fundamental difference between the two is whether the evaluation uses external knowledge or not. Internal measures are more appropriate for unsupervised tasks, particularly clustering, due to the fact that most of the relevant clustering problems do not maintain a priori information about the solution. Several internal measures employ the concept of intra/inter-cluster measures. While intra-cluster measures assess the cluster's compactness, inter-cluster measures evaluate the quality of the separation between clusters.

In two recent studies \cite{ferrari2015clustering, pimentel2019new} various internal measures were used to evaluate the performance of the clustering algorithms. A common approach for combining the individual values produced by the various measures is the \textit{average ranking}. First, for each internal measure, the performance of the clustering algorithms is ranked. Then, the average ranking position over all of the internal indices is computed for each algorithm, with the best algorithm holding the top-ranked position and the worst algorithm in the last position. 

In this paper, we use the terms 'internal measures' and 'internal indices' interchangeably. 
\subsubsection{Clustering algorithm selection based on statistical meta-features }
In the study performed by de Souto et al. \cite{de2008ranking}, the authors analyzed 32 microarray datasets that relate to cancer gene expression, using eight meta-features that are mainly descriptive statistical attributes. Seven clustering algorithms were applied to each dataset, and their performance was evaluated. Since the labels of the datasets in the study were already known (classification datasets),  the algorithms were evaluated using an external measure (and therefore, a priori information). The proposed method focused on the problem of algorithm selection for gene expression data, thus some of their proposed features reflect that domain and are not relevant to all clustering problems.

% In addition, real-world clustering tasks usually do not have a priori known solutions, thus the clustering algorithms can not be evaluated by external metrics. 
% \Noy{By analyzing datasets from various domains and 
% evaluating the algorithms by internal indices ( i.e., we do not utilize a priori information about the solution), our method 
% achieves better generalization.}
 \subsubsection{Clustering algorithm selection based on evaluation meta-features}
 Another study on gene expression analysis proposed by \cite{vukicevic2016extending} extended the statistical set of meta-features suggested in the previous work. 
The extended set of meta-features includes 19 meta-features, where five describe the characteristics of clustering algorithms, six are internal evaluation measures, and the others are the statistical features presented in \cite{de2008ranking}. So again, not all of the suggested meta-features can be applied to all clustering problems.

\subsubsection{Clustering algorithm selection based on distance meta-features}\label{distance-based}
Another study by Ferrari and de Castro \cite{ferrari2015clustering} proposed distance-based meta-features for characterizing clustering problems.
The authors built a vector \textbf{d} which contains the Euclidean distance among all instances in a dataset:
\begin{equation}
 \textbf{d}= [d_{1,2},....,d_{i,j},....,d_{n-1,n}]
\end{equation}
\noindent where $n$ denotes the number of instances, and $d_{i,j}$ defines the Euclidean distance between the $i$-th and the $j$-th instances.
 Then, by normalizing the vector in the interval [0,1], they extracted 19 meta-features from each dataset. We denote the normalized vector as $\textbf{d'}$. Table \ref{tab:distance_meta-features} describes these meta-features.
\begin{table}[ht!]
    \centering
    \begin{adjustbox}{max width=\textwidth}
    \begin{tabular}{l l}
    \hline
         Meta-feature& Description  \\
         \hline
         $MF_1$& Mean of \textbf{d'} \\
         $MF_2$& Variance of  \textbf{d'} \\
         $MF_3$& Standard of deviation of \textbf{d'} \\
         $MF_4$ & Skewness of \textbf{d'} \\
         $MF_5$& Kurtosis of \textbf{d'} \\
         $MF_6$ & \% of values in the interval [0, 0.1] \\
        $MF_7$ & \% of values in the interval (0.1, 0.2] \\
         $MF_8$& \% of values in the interval (0.2, 0.3] \\
         $MF_9$ & \% of values in the interval (0.3, 0.4] \\
        $MF_{10}$ & \% of values in the interval (0.4, 0.5] \\
        $MF_{11}$ & \% of values in the interval (0.5, 0.6] \\
        $MF_{12}$ & \% of values in the interval (0.6, 0.7] \\
        $MF_{13}$ & \% of values in the interval (0.7, 0.8] \\
        $MF_{14}$ & \% of values in the interval (0.8, 0.9] \\
        $MF_{15}$ & \% of values in the interval (0.9, 1] \\
        $MF_{16}$ & \% of values with absolute z-score in the interval [0, 1) \\
        $MF_{17}$ & \% of values with absolute z-score in the interval [1, 2) \\
        $MF_{18}$ & \% of values with absolute z-score in the interval [2, 3) \\
        $MF_{19}$ & \% of values with absolute z-score in the interval $[3, \infty)$ \\
        \hline
    \end{tabular}
    \end{adjustbox}
    \caption{Distance-based meta-features and their description.}
    \label{tab:distance_meta-features}
\end{table}
They evaluated their approach on 84 datasets from the UCI Machine Learning Repository \cite{Dua:2019} using seven clustering algorithms. The algorithms' performance was assessed based on 10 internal indices, without a priori knowledge. 

Later in the paper, we refer to this approach as the distance-based method.  
\subsubsection{Clustering algorithm selection based on correlation and distance meta-features}\label{CaD}

Recently, a study by Pimentel and de Carvalho \cite{pimentel2019new} proposed CaD,  a new approach for characterizing a dataset. Their set of meta-features combines two measures: dissimilarity and correlation. 
While the former is based on the distance between instances, the latter represents the pairwise interactions of two instances.
The distance-based meta-features were computed using the Euclidean distance among all of the instances, as described in \cite{ferrari2015clustering}. As a result, a vector \textbf{d} is created. 
Using the Spearman's rank correlation coefficient, an additional vector \textbf{c} containing the correlation among instances is generated:

\begin{equation}
 \textbf{c}= [c_{1,2},....,c_{a,b},....,c_{n-1,n}]
\end{equation}
\noindent where $n$ denotes the number of instances, and $c_{i,j}$ defines the correlation between the $i$-th and the $j$-th instances.
Next, the vectors \textbf{c} and \textbf{d} are concatenated to a new vector \textbf{m}. 
Then, by normalizing the vector \textbf{m} in the range [0,1], the meta-features described  in Table \ref{tab:distance_meta-features} are extracted.
The authors conducted a comprehensive evaluation including more clustering algorithms and a larger number of datasets than the previous studies mentioned above. 

{
\subsubsection{Algorithm selection with hyperparameter optimization}
While all the presented studies consider the algorithm selection challenge, a study by Tschechlov \cite{tschechlov2019analysis} addresses the combined problem of algorithm selection and hyperparameter optimization. This work aims to apply the concepts of supervised AutoML systems to clustering analysis. To this end, the author proposed a method that consists of two phases: offline and online. The offline phase includes a set of labeled datasets. For each dataset, meta-features are extracted, and then various configurations (algorithms with their hyperparameters) are evaluated on it. In the online phase, the goal is to provide an appropriate configuration for a new clustering task, using the knowledge obtained in the previous phase. The work focuses on partitional clustering algorithms, and as the author indicates: "This work focuses on partitional clustering algorithms, hence, it can be examined if the concept can
also be applied on other families of clustering algorithms as well. \textbf{However, this introduces several
challenges.}"
}

Despite the novelty, simplicity, and satisfactory results demonstrated in the studies described above, we believe that further enhancements are needed, and our proposed method improves upon the existing methods in three main areas.

The first area of improvement addresses the computation of multivariate measures (e.g., similarity, distance, etc.) between instances. 
High-dimensional data are now widely used in different applications. Usually, multidimensional datasets include some noisy, redundant, or uninformative features. 
% Thus, using all the features as a basis for meta-features generation can lead to a bias in the dataset characterization and as a result to a non accurate meta-model.
Thus, using all of the features as the basis for meta-feature generation can produce  misleading results.
To resolve this problem, we utilize a dimensionality reduction technique. Our method generates the meta-features based on the reduced representation of the instances so that their significant and latent essence is captured. 

Furthermore, the studies \cite{ferrari2015clustering, pimentel2019new} described above produce a meta-model based on the combination of multiple internal clustering measures. We extend this research by producing a meta-model both for multiple measures and an individual measure - a model for every internal index; the meta-model focuses on optimizing a specific clustering measure.

The characteristics of the meta-features are the last area to be considered. The meta-features in the abovementioned studies largely depend on the interactions exposed (i.e., similarity or dissimilarity) among the instances.  By constructing a similarity graph, we are able to reveal latent relationships, which can then be used as meta-features.

Graphs are widely used for modeling entity interactions in many domains, including social and biological networks \cite{goyal2018graph}.
In recent years, several convolutional neural network architectures
have been proposed to address a large class of graph-based learning problems, including learning graph representation \cite{niepert2016learning}. These architectures seek to learn a continuous vector representation $z \in$ $ \mathbb{R}^d$  for a given graph $G$. 
Since graphs naturally embody interconnectivity between data points, we propose using novel graphical embedded meta-features to address the clustering algorithm selection challenge.

Table \ref{tab:summarization_of_sudies} summarizes the main properties related to the studies reviewed and the contribution of our work toward the advancement of research in this area. 
We compare the meta-learning approaches in terms of the number of datasets used for training a meta-model, the number of clustering algorithms evaluated, the number of clustering algorithm categories (e.g., partitional algorithms, hierarchical approaches, etc.), and the number of internal measures that are applied to assess the performance of the clustering algorithm. As seen in the table, on most of these criteria, our method excels.

\begin{table}[h!]
    \centering
    \begin{adjustbox}{max width=\textwidth}
    \begin{tabular}{l c c c c c c}
    \hline
     Papers &de Souto et al. \cite{de2008ranking} & Vukicevic et al. \cite{vukicevic2016extending} & Distance-based \cite{ferrari2015clustering} & CaD \cite{pimentel2019new}  & {Tschechlov \cite{tschechlov2019analysis}}&\textbf{This study} \\
     \hline
     Number of datasets & 32& 30 &84 & 219& 81&210\\
     Number of clustering algorithms & 7 & 7 & 7& 10 & 3&{17} \\
     Number of algorithm categories  & 4 & 2&5&4 & 1&8\\
     Number of internal clustering  measures & 0 & 0& 10 &10 &4& 10\\
     \hline
    \end{tabular}
    \end{adjustbox}
    \caption{A comparison of approaches for clustering algorithm selection using meta-learning.}
    \label{tab:summarization_of_sudies}
\end{table}

%% file: Method/Method_main.tex
\section{Methods}

At the beginning of this section, we introduce a simple new popularity-based baseline that is capable of recommending suitable algorithms for a given measure without additional significant computational effort for new datasets.  Then, we present \MethodName, a novel meta-learning approach for clustering algorithm selection that achieves state-of-the-art results.   
\input{Method/Popularity-based}
\subsection{\MethodName method}
\label{Method}
\MethodName is a meta-learning framework for ranking the performance of clustering algorithms for a given  combination of dataset and clustering evaluation measure. 
\MethodName consists of two phases: training and inference. During the training phase, we evaluate and rank the clustering algorithms, analyze multiple datasets, generate meta-features, and train a ranking meta-model. During the inference phase, the  meta-model generated in the training phase is used to produce a ranked list of promising clustering algorithms for a new dataset and clustering metric.

\input{Method/sub_section_train}

\input{Method/sub_section_test}

%% file: Method/Popularity-based.tex
\subsection{Popularity-based baseline}
We suggest a new simple, yet effective, baseline method based on popularity for algorithm selection. 
A popularity-based approach is often used as a strong baseline in ranking tasks such as recommendation systems \cite{10.1145/3209978.3210014}.  

Given a set of clustering algorithms {$A$}, two steps are involved in the popularity-based baseline's preprocessing. First, for each clustering measure $m$, we count the number of times that each algorithm $a \in A$ obtained the best performance across the dataset collection (as  explained below in subsection \ref{clustering_algorithms_evaluation}). We denote this score as $a_{m\_\#top}$.  Then, a rank is assigned to every algorithm based on its score $a_{m\_\#top}$. 
As a result,  for each clustering index $m$, a ranked vector $W_{m}$ represents the popularity of the algorithms generated. The algorithms are ranked in ascending order, such that the most frequent algorithm (i.e., the algorithm that most frequently obtained the best performance) holds the first rank position.

Then, given a previously unseen dataset $d_{new}$ and a measure $m$ selected by the user, the popularity-based approach provides the vector $W_{m}$. The vector $W_{m}$ is precalculated, and therefore no additional computational effort is required for a new dataset.  

 To illustrate how the popularity-based baseline works, consider a case in which the following algorithms are examined: an evolutionary algorithm for clustering (EAC), particle swarm optimization for clustering (PSC), minimum spanning tree (MST), agglomerative single-linkage (SL), agglomerative average-linkage (AL), agglomerative complete-linkage (CL), agglomerative Ward-linkage (WL), K-means (KM), K-harmonic means (KHM), kernel K-means (KKM), mini-batch K-means (MBK), fuzzy C-means (FC),  density-based spatial clustering applications with noise (DBSCAN), {mean shift (MS), a Gaussian mixture model with a full matrix (GMF), a Gaussian mixture model with a tied matrix (GMT), and  a Gaussian mixture model with a diagonal matrix (GMD)}; these algorithms and the internal clustering measures are described in more detail in subsections \ref{clustering_algorithm} and \ref{clustering_evaluation_measure}, respectively. Table \ref{tab:algorithms} shows the number of times each algorithm held the first rank position.  For example, if the selected measure is the Dunn index, {then the DBSCAN  algorithm gets the first rank position, followed by SL, AL, EAC, CL, MST, KM, KKM, KHM, GMF, MS, PSC, WL, FC, GMT, and GMD, respectively}.
 Table \ref{tab:top-3performin } shows the three top-performing algorithms for each internal clustering measure.
% reasonable results without computational effort. 

\begin{table}[ht!]
    \centering
     \begin{adjustbox}{max width=\textwidth}
    \begin{tabular}{ l c c c c c c c c c c c c c c c c c}
    \hline
        Index & EAC & PSC& MST&SL&AL&CL&WL&KM&KHM&KKM&MBK&FC&DBSCAN &MS &GMF& GMT & GMD \\
        \hline
        
         Bezdek-Pal & 19 &0&10&\textbf{117}&29&5&3&3&2&5&0&2&14&1&0&0&0\\
         Dunn Index & 26&1&12&38&34&22&1&8&6&7&3&1&\textbf{44}&2&3&1&1 \\
         Calinski-Harabasz& 6&4&1&1&4&4&12&\textbf{78}&18&31&7&26&1&1&4&5&7\\
         Silhouette Score & 37&1&3&25&\textbf{40}&9&18&35&4&5&4&4&4&3&5&4&9\\
         Milligan-Cooper &   27&6&6&28&\textbf{44}&22&5&16&3&11&5&1&24&2&3&6&1\\
         Davies-Bouldin& 8&0&3&\textbf{108}&52&5&5&9&2&0&2&3&1&2&1&3&6\\
         Handl-Knowles-Kell& 18&4&26&\textbf{85}&27&5&7&12&3&3&2&3&9&0&3&3&0\\
         Hubert-Levin& 29&7&2&41&\textbf{43}&22&9&24&2&0&5&3&8&2&6& 1&6\\
         SD-Scat&11&10&3&\textbf{70}&33&9&4&11&15&0&3&5&17&3 &4&3&9\\
         Xie-Beni& 5&0&11&\textbf{67}&37&22&11&3&1&1&3&8&34&3&0&3&1\\
         \hline
         Average ranking&28&0&2&35&\textbf{67}&18&17&23&2&1&2&5&0&3&0&4&3\\
         \hline
    \end{tabular}
    \end{adjustbox}
   \caption{Number of times that each algorithm achieved the highest performance for each of the  internal clustering measures, over 210 datasets.  The most frequent algorithm for each internal measure is highlighted.}
    \label{tab:algorithms}
\end{table}

\begin{table}[h!]
    \centering
    \begin{tabular}{ l c c c }
    \hline
        Index & Top-1 & Top-2& Top-3 \\
        \hline
        
         Bezdek-Pal & SL& AL& EAC\\
         Dunn Index & DBSCAN & SL& AL \\
         Calinski-Harabasz& KM& KKM& FC \\
         Silhouette Score &  AL& EAC& KM\\
         Milligan-Cooper & AL& SL&EAC  \\
         Davies-Bouldin& SL & AL & KM\\
         Handl-Knowles-Kell& SL&{AL}& {MST}\\
         Hubert-Levin& {AL} & {SL} &{EAC}\\
         SD-Scat&{SL}&AL&{DBSCAN}\\
         Xie-Beni& SL& AL & {DBSCAN}\\
         \hline
         Average ranking&AL &{SL} &{EAC}\\
         \hline
    \end{tabular}
   \caption{
  The three top-performing algorithms for the various   internal clustering measures.}
    \label{tab:top-3performin }
\end{table}

%% file: Method/sub_section_train.tex
\subsection{The training phase}
\begin{figure}[h!]
    \centering
     \begin{adjustbox}{max width=\textwidth}
    \includegraphics{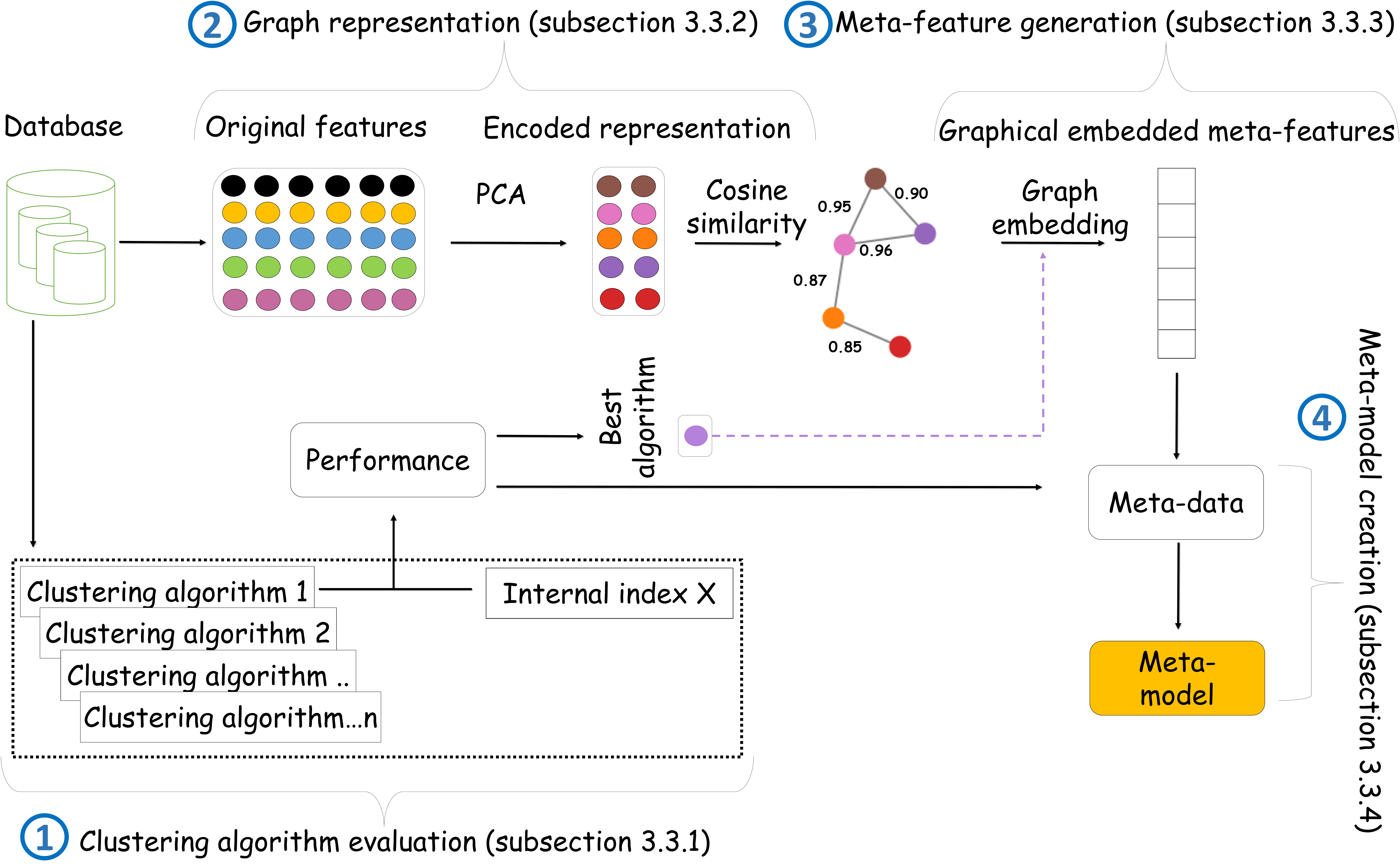}
    \end{adjustbox}
    \caption{The training phase -  The \MethodName workflow starts with a collection of datasets. The clustering algorithms are then evaluated on the datasets using an internal measure (step \#1). In the graph representation step the PCA algorithm is applied, and a weighted graph is constructed (step \#2). Then, the graph is transformed into a lower-dimensional space (step \#3). Finally, a meta-model capable of recommending high-performing algorithms for  previously unseen datasets is created (step \#4). }
    \label{fig:workflow}
    
\end{figure}

During the training phase, we evaluate clustering algorithms on a large collection of diverse datasets. Then, we generate meta-features that are used to train a meta-model.
% It is important to highlight that an unique meta-model is designed for each of the following internal indices that are described later. 
The training phase includes four steps and is illustrated in Figure \ref{fig:workflow}, which is followed by a description of each step.

\input{Method/Algorithms_evaluation}

\input{Method/graph_representation}

\input{Method/meta-features_generation}
\input{Method/meta-model_creation}

%% file: Method/Algorithms_evaluation.tex
\subsubsection{Clustering algorithm evaluation}\label{clustering_algorithms_evaluation}
In this step (step one), in order to quantify the quality of the obtained solution for each clustering algorithm on every dataset, the following process is used: 
\begin{enumerate}
    \item Given a collection of datasets $D$, a set of clustering algorithms {$A$}, and an internal index $m$, we quantify the performance of all combinations of $d \in D$ and { $a \in A$} using $m$. We denote the result of this evaluation as {$P_{a,d,m}$}. Based on the performance, we rank the clustering algorithms for each dataset, such that the algorithm with the best performance occupies the first rank position, and the algorithm with the worst performance holds the last position. The ranking of each algorithm is defined as {$R_{a,d,m}$}.
    \item For each $d \in D$, we select the algorithm with the best performance for the index $m$.  We refer to this value as {$a_{best,d,m}$}, namely $R_{a_{best},d,m} = 1$.
    \item  We generate the pair { $(d, a_{best,d,m})$} and consider the second entry as a label for dataset $d$.
\end{enumerate}

With regard to step one, it is also important to mentioning the following two points.
First, to deal with the non-deterministic nature of some algorithms in {$A$}, we repeat this assessment 10 times and compute the average results; thus, the result {$P_{a,d,m}$} is computed based on  the average performance score of algorithm $a$ on $d$, using $m$.
    % \item \MethodName enables two possibilities for evaluating the performance of the clustering algorithm by the internal index drawn in Figure \ref{fig:workflow}: (1) using individual metric, (2) average ranking of the performance achieved by multiple indices.  
    Second, the  individual  internal  index  mentioned  in  Figure \ref{fig:workflow}  can  be  replaced  by computing the average ranking score of each clustering algorithm's performance based on multiple internal measures  (see subsection \ref{clustering_performance_assessment} for more details).

%% file: Method/graph_representation.tex
\subsubsection{Graph representation} \label{graph_generation}
The principal goal of step two is to generate graphical representations of the datasets. We first describe our motivation, and then we demonstrate the entire process.
 
Our study is based on the intuition that similar types of datasets are likely to induce similar performance from clustering algorithms, thus if a given clustering algorithm is effective for a dataset it will likely also be well-suited for similar datasets.
% thus they can be solved by the same algorithm. 
We further expect that the performance of the given clustering algorithm {$a$} on dataset $d$ (denoted by {$p(d,a)$})  is mainly influenced by the pairwise connections between the instances of a dataset.  Thus, by modeling these interactions in a graph and creating embedding representations (i.e., our meta-features) that capture their essence, we will be able to infer dataset similarity in terms of the algorithm performance. Formally, given a function $g$, which receives a dataset as input and returns its embedding, we assume that if $g(d_1) \approx g(d_2)$, then $p(g(d_1),a) \approx p(g((d_2),a))$.     

As known, given a set of data points $x_1,...x_n$ and pairwise similarities, clustering algorithms seek to expose the intrinsic sub-classes in the data by grouping the data into clusters so that data points within a cluster are more similar to each other than to 
those outside the cluster. 

Since graphs naturally embody the notions of local neighborhood interactions as well as global properties, the translation of the datasets into graphs is straightforward. 
Furthermore, converting the datasets into graphs is motivated by the fact that each dataset has its own dimensions, and thus there is no common denominator across the different datasets. The graphical representation enables a unified representation of the various datasets.

Applying a graph-based method to extract informative meta-features that can quantify dataset similarity enables us to recommend a suitable algorithm for a new dataset. 

To obtain the representation of the dataset's interactions, we adopt the following procedure:

First, each dataset undergoes preprocessing that includes the following steps, as done in previous research \cite{ferrari2015clustering}:

(a) Removing the class/label column

(b) Encoding the nominal attributes 
into numeric values

(c) Removing features with identical values for all of the instances

(d) Removing features with a distinct value for each instance

(e) Removing features with more than 40\% NaN values

(f) Normalizing all features in the range [0,1]

{The features described in steps (c) and (d) were removed, since they do not help distinguish between two different instances.}

Then, we apply a dimensionality reduction technique. Dimensionality reduction is the mapping of high-dimensional data into a meaningful reduced representation \cite{garcia2015data}.
In this paper, we apply the principal components analysis (PCA) method \cite{dunteman1989principal} with low information loss. One of the most important linear dimensionality reduction techniques, PCA is defined as an orthogonal linear transformation that converts the data into a lower-dimensional space by identifying the dimensions with the greatest variance. 

Next, inspired by \cite{von2007tutorial}, we construct a \textit{similarity graph} $G=(V,E)$ to represent the local characteristics of the data. 
A similarity graph is an undirected and weighted graph based on the cosine similarity measure.  In a similarity graph, 
each vertex $v_i \in V$ corresponds to an instance $x_i$, and two vertices are connected via an edge if the similarity $s_{i,j}$ between the corresponding instances $x_i$ and $x_j$ is over a certain threshold (we set the threshold value at 0.9). The computation of $s_{i,j}$ is based on the reduced representation of $x_i$ and $x_j$, and the result determines the weight of the edge $e=(i,j) \in E$.

%% file: Method/meta-features_generation.tex
\subsubsection{Meta-feature generation}\label{meta-features_generation}
During step three, given a graph $G$, we utilize a graph convolutional neural network (GCNN) technique  to generate {an embedding representation for the entire graph. } Motivated by the success of convolutional neural networks (CNNs) in computer vision, numerous studies %interest 
extended those methods to graph data \cite{kipf2016semi}. In this study, we incorporate both the graph structure and the graph label into an embedding representation using a supervised GCNN model.
We stack multiple graph convolutional layers with a readout layer so that the graph convolutional layers are responsible for extracting high-level node representations, and the readout layer aggregates node representations into a graph representation.  This graph representation is the desired embedding representation, i.e, the meta-features. The graph representation is then fed into a classifier to predict the graph's label. 
% Then, this representation is fed into a classifier to predict the graph label. 

\textbf{Definition}: Given a set of labeled graphs $GL=\{(G_1,l_1),...,(G_n,l_n)\}$, our goal is to learn a function $f: G_i \rightarrow \mathbb{R}^d$, such that $ max(softmax(f(G_i))) =l_i \in L$, and $L$ is the set of graphs labels. 

Each graph {$G_i$=$(V,Z,B,X)$} is comprised of:
(1) a set $V$ consisting of nodes $\{v_1,\dots,v_n\}$,
(2) an adjacency matrix {$Z$} $\in \mathbb{R}^{n\times n}$ where {$z$}$_{ij}$ denotes the weight of the edge between the vertices $v_i$ and $v_j$, (3) a degree matrix {$B$}, and (4) a node feature vector matrix $X \in \mathbb{R}^{n \times d}$. 
Each graph $G_i$ also has a corresponding label $l_i$.

According to a study by Kipf and Welling \cite{kipf2016semi}, each graph convolution layer is defined as follows:
\begin{equation}
    H^{(l+1)} = \sigma(FH^{(l)}W^{(l)})
\end{equation}

\noindent {where $F= \tilde{B}^{-\frac{1}{2}}\tilde{Z}\tilde{B}^{-\frac{1}{2}}$} is the normalized adjacency matrix with added self-loops,  i.e., $\tilde{Z}= Z + I_n$, $I_n$ is the identity matrix;  $\tilde{B}_{ii} = \sum_{j}{\tilde{Z}_{ij}}$, $W^{(l)}$ is a layer-specific weight matrix; $\sigma(\cdotp)$ is an activation function (e.g., ReLU); and $H^{(l)}$ denotes the node representation matrix in the $l^{th}$ layer. Finally, $H^{(0)} =X$.

Here, the set of graph labels $L$ consists of the clustering algorithms described in subsection \ref{clustering_algorithm}, and therefore $L = A$. Now, given an internal measure {$m$}, the corresponding label for dataset  $d_j \in D$, which is modeled by $G_j$, is {$a_{best,d_j, m}$} (see subsection \ref{clustering_algorithms_evaluation}).
In addition, $X$ is defined using the DeepWalk framework for weighted graphs.
DeepWalk is an effective empirical method for learning a fixed-size embedding representation  \cite{perozzi2014deepwalk} for each vertex in a graph. 
Given a graph $G=(V,E)$, DeepWalk generates the matrix  $X \in \mathbb{R}^{|V| \times d}$, { where $d$ is a small number of latent dimensions}. 
 The method's ability to capture the graph topology and the fact that $d$ is a predetermined parameter, allow us to create an enhanced GCNN model with a  fixed-size feature dimensions. 

Algorithm \ref{Alg:generation} summarizes the \textit{meta-feature generation} step.
\input{Algorithms/generation.tex}

%% file: Algorithms/generation.tex
\setcounter{algocf}{0}
\begin{algorithm}[h!]
\SetAlgoLined
\DontPrintSemicolon
\KwInput{$m$ - an internal clustering measure,  $GL$ = $\{(G_1 ,a_{best,d_1, m}), \dots, (G_n, a_{best,d_n, m})\}$ - a set of labeled graphs}
  \KwOutput{$Meta\_features\_set$ - The set of meta-features }

    \SetKwFunction{FMain}{MFG}
    \SetKwProg{Fn}{Function}{:}{}
    \Fn{\FMain{$m$, $GL$}}{
    $ {X\_matrices} \longleftarrow \emptyset $\\
     \ForEach{{$<G_i, a_{best,d_i, m}> \in GL$}} {
     $ X_i   \longleftarrow WeightedDeepWalk{(G_i)}$ \\
     $ X\_matrices   \longleftarrow X\_matrices \cup X_i$ \\

    }
    $ GCNN\_model \longleftarrow CreateGCNN\_model(GL, X\_matrices)  $\\
    $Meta\_features\_set \longleftarrow GCNN\_model.getAllEmbeddings()$ \\

\textbf{return} $Meta\_features\_set$ 
}
\textbf{End Function}
\caption{Meta-feature generation(MFG)}
\label{Alg:generation}
\end{algorithm}
\normalsize

%% file: Method/meta-model_creation.tex
\vspace{-4mm}
\subsubsection{Meta-model creation}
The key idea of this work is to produce an enriched meta-model (step 4) capable of recommending the top-performing clustering algorithm(s) for an unseen dataset and clustering evaluation measure.

To train a ranking model that optimizes a clustering evaluation metric $m$, we collected a large labeled training set. For each combination of  $d \in D$ and  and { $a \in A$} (as previously mentioned, $D$ and $A$ are respectively collections of datasets and clustering algorithms), we create a training instance t, where eventually, $t = \{M_{d, m} \cup M_{a} \cup R_{a,d, m} \}$. Training instance $t$ includes: 
\begin{enumerate}
    \item $M_{d, m}$ - the set of meta-features generated in the \textit{meta-feature generation} step (explained in subsection \ref{meta-features_generation})
    \item $M_{a}$ - a single discrete feature describing $a$
    \item $R_{a,d, m}$ - the rank obtained by algorithm  $a$ for dataset $d$ using $m$
(see subsection \ref{clustering_algorithms_evaluation} for more details)
\end{enumerate}

% Eventually, $t = \{M_{d, m} \cup M_{a} \cup R_{a,d, m} \}$.

Next, we use the ranking version of the XGBoost algorithm as a meta-learner, since prior research  \cite{cohen2019autogrd} showed that XGBoost is well suited for producing a list of promising candidates.
Once the meta-model has been trained, predictions for previously unseen datasets can be made.
% The \textit{meta-model creation} step is summarized in Algorithm \input{Algorithms/meta-model_creation} \ref{alg:model_creation}.
The \textit{meta-model creation} step is summarized in Algorithm \ref{Alg:metamodel_creation}.
\input{Algorithms/meta-model_creation}
% \begin{figure}[h!]
%     \centering
%     \includegraphics{Figures/Algorithm2.pdf}
%     \label{fig:my_label}
% \end{figure}

%% file: Algorithms/meta-model_creation.tex
% \begin{algorithm}[h!]
% \caption{Meta-model creation}\label{alg:model_creation}
% \begin{algorithmic}[1]
% \Procedure{CreateModel}{\raggedright Measure $m_i$, GL=$\{(G_1 ,R_{d_1\_best,m_i}), \dots, (G_n, R_{d_n\_best,m_i})\}$, SS =$\{S_1, \dots, S_n\}$, algorithms $L$, datasets $D$ }
% \State  $ MetaFeatures\_set \gets \textbf{MPG}( m_i, GL, SS)$ \Comment{See Algorithm 1}
% \State $ training\_set \gets \emptyset $
% \For{\textbf{each} $d_j$ in D}
% \State $ M_{d_j, m_i} \gets MetaFeatures\_set.\textbf{GetMetaFeatures}(d_j) $
% \State $ P_{d_j,m_i} \gets \textbf{EvaluatePerformance}(d_j, L, m_i)  $
% \For {\textbf{each} $l_a$ in L}
% \State $ M_l_a \gets \textbf{GetDiscreteFeature}(l_a) $
% \State $ R_{l_a,d_j,m_i} \gets P_{d_j, m_i}.\textbf{GetRanking}(l_a)  $
% \State $ current\_features \gets (M_{d_j, m_i} \cup M_l_a \cup R_{l_a,d_j,m_i} ) $
% \State $ training\_set \gets (training\_set \cup current\_features) $
% \EndFor
% \EndFor
% \State $ RankingMetaModel \gets \textbf{XGBoost}(training\_set)  $ 
% \State $ \textbf{return RankingMetaModel}  $
% \EndProcedure
% \end{algorithmic}
% \end{algorithm}
\setcounter{algocf}{1}
\begin{algorithm}[h!]
\SetAlgoLined
\DontPrintSemicolon
\KwInput{$m$ - an internal clustering measure,  $GL$ = $\{(G_1 ,a_{best,d_1, m}), \dots, (G_n, a_{best,d_n, m})\}$ - a set of labeled graphs, {$A$ - a set of clustering algorithms}, $D$ - a set of datasets}
  \KwOutput{$Ranking\_meta\_model$ - The ranking meta-model}

    \SetKwFunction{FMain}{MetaModelCreation}
    \SetKwProg{Fn}{Function}{:}{}
    \Fn{\FMain{$m$, $GL$, $A$, $D$}}
    {
    $ Meta\_features\_set \longleftarrow MPG(m, GL)$ \\
    $training\_set \longleftarrow \emptyset$\\
     \ForEach{$d \in D$} {
     $ M_{d,m}     \longleftarrow Meta\_features\_set.GetMetaFeatures(d)$ \\
     $ All\_performances   \longleftarrow EvaluatePerformance(d,A,m)$ \\
    \ForEach{$a \in A$}{
    $M_{a} \longleftarrow GetDiscreteFeature(a)$\\
    $R_{a,d,m} \longleftarrow All\_performances.GetRanking(a)$\\
    $current\_features \longleftarrow (M_{d,m} \cup M_{a} \cup R_{a,d,m})$\\
    $training\_set \longleftarrow (training\_set \cup current\_features)$ \\

    }
    
    }
    $Ranking\_meta\_model \longleftarrow XGBoost(training\_set)$ \\

\textbf{return} $Ranking\_meta\_model$ 
}
\textbf{End Function}
\caption{Meta-model creation}
\label{Alg:metamodel_creation}
\end{algorithm}
\normalsize

%% file: Method/sub_section_test.tex
\subsection{The inference phase}
\begin{figure}[h!]
    \centering
     \begin{adjustbox}{max width=\textwidth}
    \includegraphics{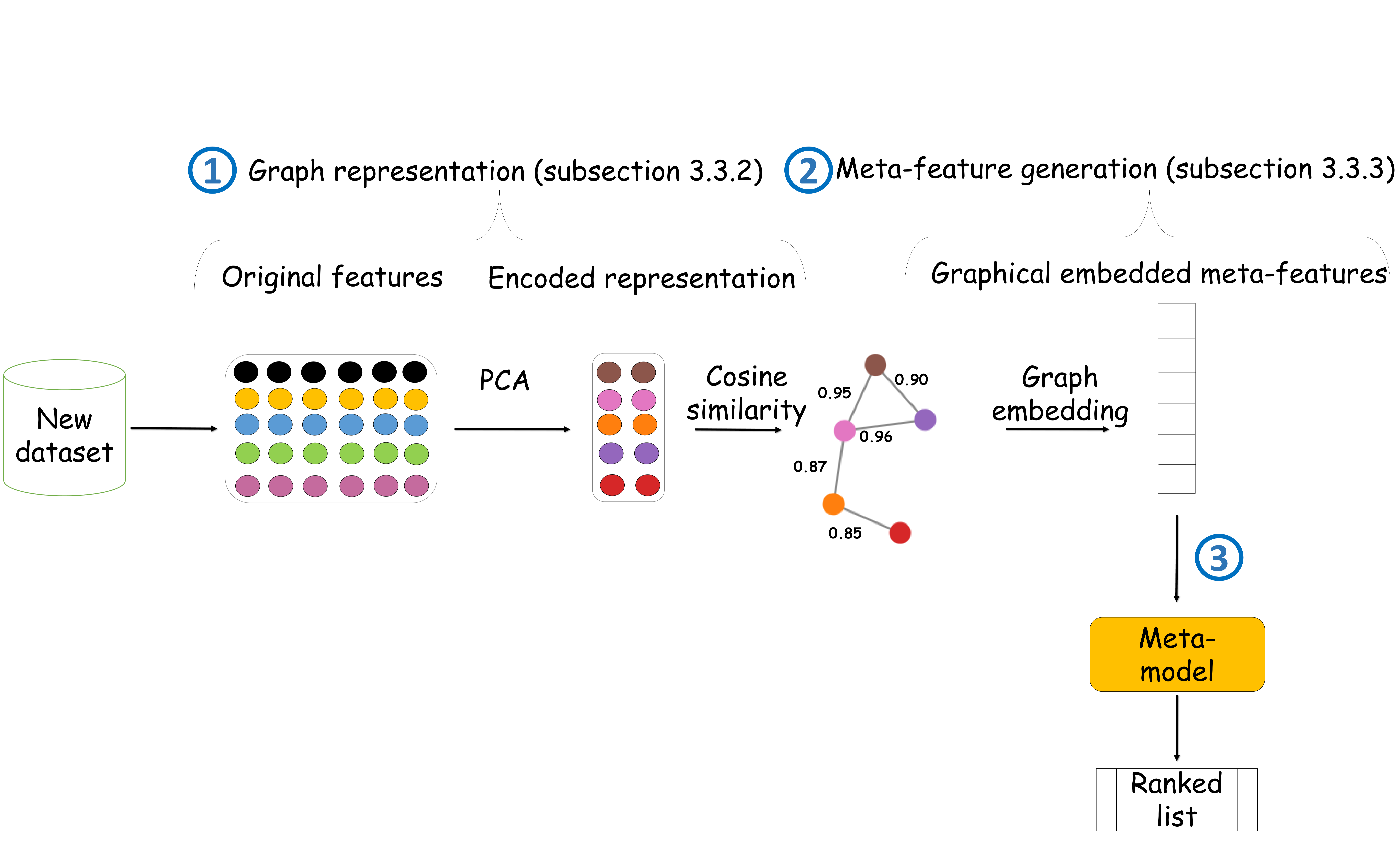}
    \end{adjustbox}
    \caption{ The inference phase - given the meta-features of a previously unseen dataset, \MethodName selects a trained meta-model according to the clustering index chosen by the user. Then, \MethodName produces a ranked list of clustering algorithms. }
    \label{fig:Inference}
    
\end{figure}
During the inference phase, we select a trained meta-model based on the  clustering measure chosen by the user and generate a list of clustering algorithms, sorted by their predicted performance on a new dataset $d_{test}$.

% During the inference phase, we generate
% a list of clustering algorithms, sorted by their predicted performance on a new dataset $d_{test}$. Given trained meta-models for each clustering internal measure, we select the appropriate meta-model according to the preferred measure by the user.
% During the inference phase, in order to generate a list of clustering algorithms to a new dataset $d_{test}$ given trained meta-models for each clustering internal measure, we select the appropriate meta-model according to the prefered measure by the user.  
% During the inference phase, we utilize the trained meta-model to generate a list of clustering algorithms, sorted by their predicted performance on a new dataset $d_{test}$.
This phase is described in Figure \ref{fig:Inference} and consists of three steps:
\begin{enumerate}
    \item We generate a graph $G_{test}$ which models $d_{test}$ by following the process described in subsection \ref{graph_generation}.
    \item  Given $G_{test}$, a set of meta-features $M_{test}$ are created (see subsection \ref{meta-features_generation}). 
    \item For each $a \in A$, we generate $M_{a}$ (i.e., a nominal feature describing $l$) and create the joint meta-feature vector $\{M_{test} \cup M_{a}\}$.
Then, based on all of the meta-feature vectors, the XGBoost trained meta-model produces a list of all clustering algorithms in $A$, ranked by their predicted performance.
\end{enumerate}

%% file: Experiments/Experiments_average.tex
\section{Evaluation}
To assess our method as a meta-learning approach that addresses the task of recommending the appropriate algorithm for a given dataset and evaluation measure, two different modes were evaluated: the average ranking model mode and the individual model mode.
As described below, we conducted an extensive evaluation with 210 datasets, {17} clustering algorithms, and 10 internal clustering measures. The two state-of-the-art methods \cite{pimentel2019new} are compared to \MethodName using the following measures: mean reciprocal rank and Spearman’s rank correlation coefficient.

\subsection{Evaluation modes}
We evaluated the methods with two different modes:
\begin{enumerate}
    \item \textbf{Average ranking model mode - } In this mode, 10 diverse  internal clustering measures are combined to evaluate the clustering algorithms' performance on the different datasets. For each dataset and clustering algorithm, 10 different estimations were produced (by the 10 different indices). To merge them,  we used the \textit{average ranking} combination technique which has been used in previous studies \cite{ferrari2015clustering, pimentel2019new}. First, for each internal measure, the performance of the {17} clustering algorithms is ranked.  Then, the mean position value for each algorithm on all indices is computed. As a result, the best algorithm is in the top-ranked position, and the worst algorithm is in the last position.
    \item \textbf{Individual model mode -} In this mode, the solutions of the clustering algorithms are ranked by a particular internal measure.
% Based on the achieved rankings,  ten unique ranking meta-models were built separately  for each of the ten internal measures,  where each of them optimizes certain clustering measure. 
A unique ranking meta-model is built for each of the 10 internal indices.
In total, 10 different meta-models were built, each of which optimizes a specific clustering measure. 
\end{enumerate}
\subsection{Datasets}
\begin{figure}[h!]
    \centering
    \begin{adjustbox}{max width=0.75\textwidth}
    \includegraphics{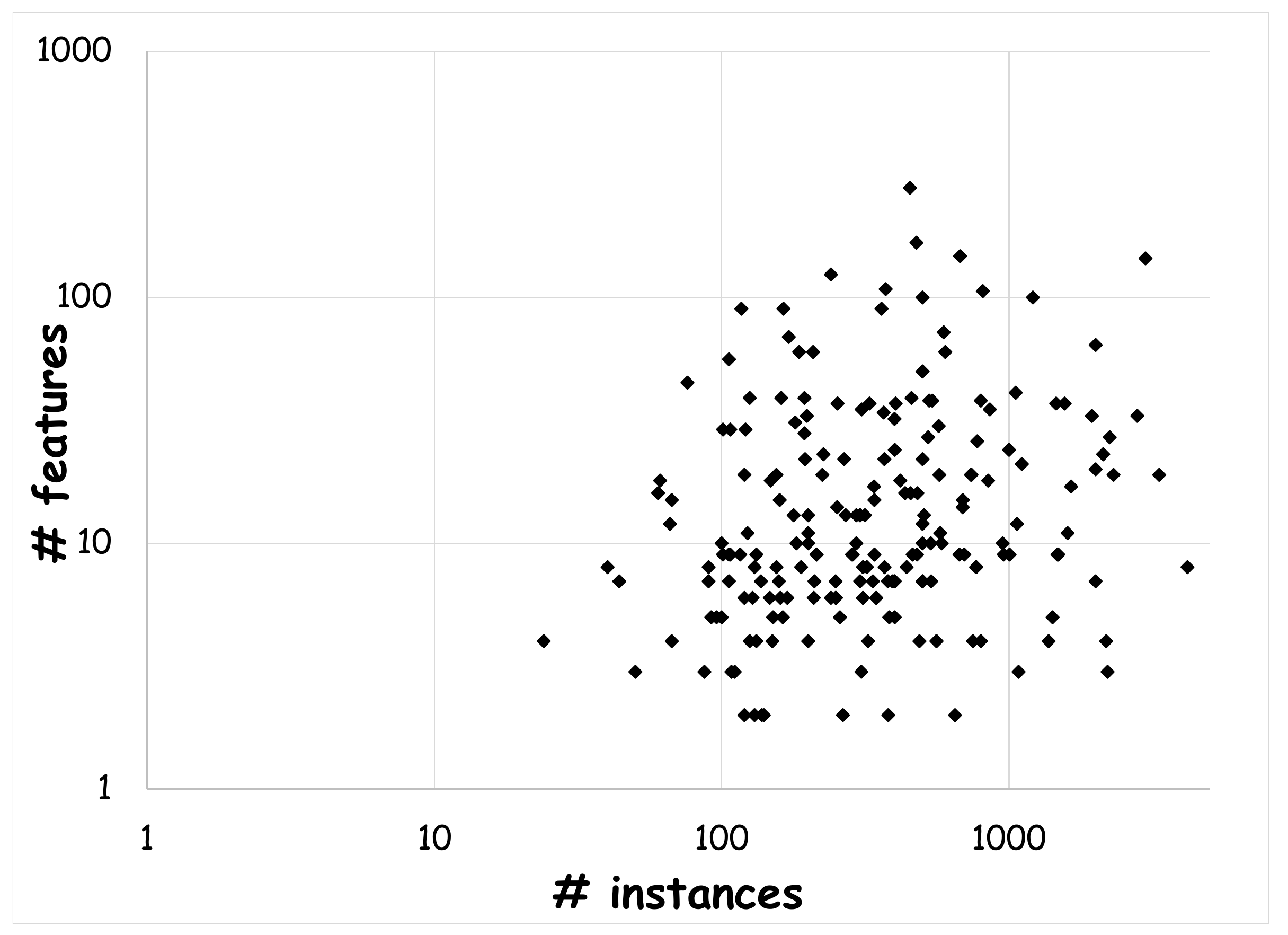}
    \end{adjustbox}
    \caption{Dimensions of the datasets used in this study.}
    \label{fig:Datasets}
\end{figure}

We conduct our experiments on 210 datasets,\footnote{All datasets are merged in https://bit.ly/3o050op} all of which are available in the following online repositories: Kaggle,\footnote{https://www.kaggle.com} KEEL,\footnote{https://sci2s.ugr.es/keel/datasets.php} OpenML,\footnote{https://www.openml.org} and UCI.\footnote{https://archive.ics.uci.edu/ml/datasets.php} 
These datasets cover a wide range of domains (e.g., medicine, natural science, etc.) and are quite varied with respect to their size, the number of features, and their composition. Figure \ref{fig:Datasets} shows the dimensions of the datasets.

Three sets of meta-features were extracted from each dataset:
\textbf{(a)} distance-based features \cite{ferrari2015clustering}, \textbf{(b)} correlation and distance-based features, denoted as CaD \cite{pimentel2019new}, and \textbf{(c)} graphical embedded-based features. (The first two, which were used in prior studies, are described in subsections \ref{distance-based} and \ref{CaD}, respectively, while the latter, which is proposed in this study, is described in subsection \ref{Method}.)

\subsection{Measures}
Since our aim is to recommend high-performing algorithms for a given combination of dataset and evaluation measure, we use the mean reciprocal rank (MRR) measure, which is a statistical measure used to evaluate the rank of the first correct recommendation:
\begin{equation}
    MRR = \frac{1}{|D|}\sum_{i \in D}\frac{1}{rank_i}
\end{equation}
where $D$ refers to the number of datasets evaluated, and ${rank_i}$ represents the ranking of the algorithm with the top performance for the $i-th$ dataset. The MRR assumes values over the range [0,1], where one indicates that the top-performing algorithm was recommended.
By using this measure, we can determine the average ranking of the first correct recommendation. 
% To the best of our knowledge, previous related works did not assess their results with the MRR measure.

Additionally, in order to estimate the quality of the recommended algorithm list, the Spearman's rank correlation coefficient (SRC) is used.
Given the actual ranking $a=(a_1,\dots, a_n)$ and the predicted ranking $p=(p_1,\dots ,p_n)$, the SRC is used to evaluate the similarity between these ranked lists. The SRC measure is defined as follows:

\begin{equation}
    SRC(a,p) = 1 - \frac{6\sum_{i=1}^n (a_i - p_i)^2}{n^3 -n}
\end{equation}
where $n$ is the number of candidate algorithms. The SRC value ranges from [-1,1]. The larger the value, the greater the similarity between the actual and predicted ranking.  This measure has been used in previous meta-learning studies \cite{de2008ranking, ferrari2015clustering, pimentel2019new} for assessing the degree of agreement between the actual and recommended lists.

\input{Method/clustering_algorithms}

\input{Method/Internal_indices}

\subsection{Experimental setup}\label{setup}
We used the following setup in all of our experiments:
\begin{enumerate}
    \item We used DeepWalk's default parameters for a weighted graph to create the feature matrix $X$ described in subsection \ref{meta-features_generation} (number of walks = 10, representation size = 64, walk length = 40, window size = 5).
    \item GCNN models:
    \begin{itemize}
        \item We used the Deep Graph Library (DGL) \cite{wang2019deep} to build the GCNN models (average ranking model and the 10 individual models for each internal measure). 
        \item {The GCNN model is based on the model proposed in the study by Kipf and Welling \cite{kipf2016semi}. As shown in Figure \ref{fig:GCNN}, four graph convolutional layers were applied sequentially with ReLU as an activation function, followed by a readout layer. While the graph convolutional layers extract high-level node representations, the readout layer collapses the  node representations obtained into a single graph representation. Every model consists of four graph convolution layers.   The graph representation is then fed into a classifier with one linear layer to obtain pre-softmax logits. The input for training the GCNN model is a set of graphs, where each graph $G_i$ has: (1)  label  $a_{best,d_i,m}$,  the algorithm that obtained the best performance on $d_i$, and (2) its node feature matrix $X_i$ (obtained by DeepWalk).} 
         \item We train all GCNN models for a maximum of 60 epochs (training iterations) using the Adam optimizer with a learning rate of 0.006. We stop training if the loss does not decrease for 10 consecutive epochs. 
        \item  We set the embedding dimensions for each dataset in every model to 300. 
        \item The values of the learning rate, the number of graph convolutional layers, and embedding size were determined in a preliminary experiment and were fixed for all datasets.\footnote{Learning rate of \{0.001, 0.002, 0.003, 0.004, 0.005, 0.006, 0.01\}, number of graph convolutional layers \{2, 3, 4, 5, 6\}, and embedding dimensions of \{50, 100, 200, 300, 400, 500\} were tested, and respectively 0.006, 4, 300 were found to produce the best results with reasonable efficiency across all models.} 
        \begin{figure}[h!]
            \centering
            \begin{adjustbox}{max width=\textwidth} \includegraphics[]{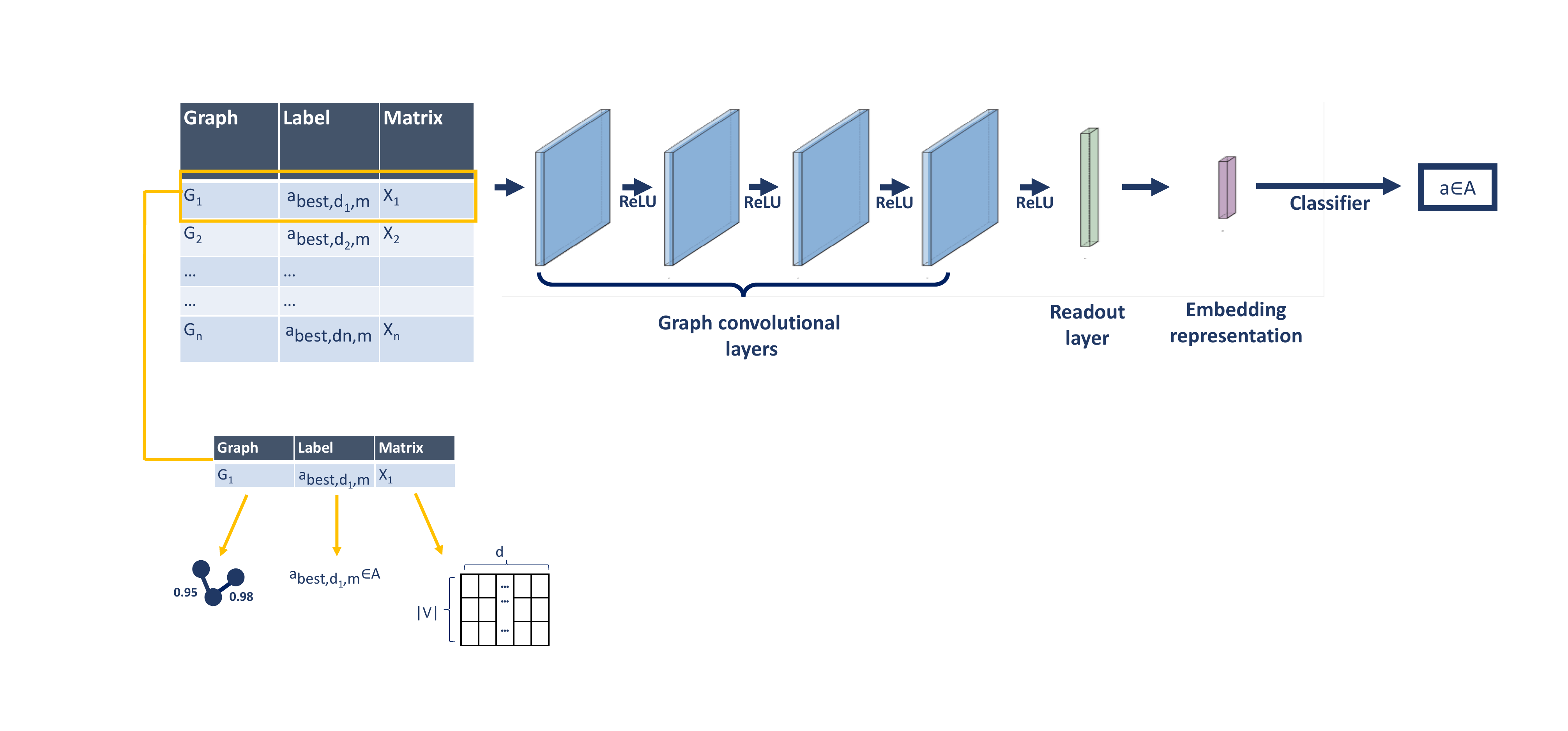}
             \end{adjustbox}
            \caption{{The GCNN architecture. $A$ is a set of clustering algorithms.}}
            \label{fig:GCNN}
        \end{figure}
       
    \end{itemize}
    
    \item We evaluated the performance of all of the meta-learner models using the leave-one-out procedure. For each dataset evaluated, $d_i \in D$, we trained the ranking model using meta-features from $d_j\in D$, where $i \neq j$, and D is a collection of datasets. Then, $d_i$ is used to test the XGBoost model generated.
    {   We evaluated MARCO-GE using the leave-one-out cross-validation technique. Consequently, the reported results were calculated by averaging the evaluation measure values for all of the datasets tested. Since each dataset is used once as a test, the mean value is computed based on 210 (the total number of datasets) estimators.   }
    % \item We used the XGBoost algorithm to train the algorithm ranking
    % model. We chose the objective function: ’rank:pairwise’ and
    % set the algorithm’s parameters empirically using the leave-one-out
    % approach.  The parameters of each model will be available in the code repository. 
    \item We used the XGBoost algorithm to train the ranking
    model. We chose the objective function: ’rank:pairwise.’ For each clustering measure, we set the hyperparameters based on a preliminary study. These parameters are identical for all of the datasets used and all of the methods evaluated. The hyperparameters of each model are available in the code repository. 
    
\end{enumerate}

\section{Evaluation results}
In this section, we first present the results and an analysis of the average ranking mode. Then, we perform  the individual model mode evaluation and discuss the results of the popularity-based baseline.  Next, we analyze both the importance of applying the PCA algorithm and \MethodName's sensitivity across several hyperparameters. {Finally, we investigate  the computational complexity and computational cost  of \MethodName.}
\subsection {Evaluation results: average ranking model mode}
The three meta-learning methods: distance-based, CaD, and \MethodName are compared with each other and with the \textit{standard ranking} baseline. The \textit{standard ranking} baseline represents the average ranking over all of the training datasets. The results of our evaluation are shown in Table \ref{tab:average_ranking_table}, where it is clear that \MethodName has both the highest average SRC and the highest average MRR over the 210 datasets evaluated. 
\begin{table}[h!]
    \centering
    \begin{tabular}{c c c c c}
    \hline
        Measure &Standard Ranking & Distance-Based & CaD & \MethodName  \\
        \hline
         SRC&0.627 & 0.631& 0.629 & \textbf{0.645} \\
         MRR& 0.498& 0.509& 0.499 & \textbf{0.822}\\
         \hline
    \end{tabular}
    \caption{Average ranking model results - the average SRC and MRR values of the methods evaluated over 210 clustering datasets. The best results with statistical significance are highlighted.}
    \label{tab:average_ranking_table}
\end{table}
As shown in Table \ref{tab:average_ranking_table}, \MethodName also outperforms the standard ranking baseline by a wide margin. 

Not only did \MethodName suggest the most similar recommendation (with regard to the actual ranking) with an average SRC of 0.645, it also needs to sample much fewer algorithms in order to make a perfect recommendation.
% The results of our method not only suggest the most similar recommendation to the actual ranking, but also indicate that \MethodName needs to sample much fewer algorithms in order to achieve perfect recommendation.

We used the Friedman test to validate the statistical significance of the differences between the methods evaluated  \cite{demvsar2006statistical} for each measure separately.  The null hypothesis that the four methods perform the same and the observed differences are merely random was rejected with $p$ \textless 0.01. We proceeded with Wilcoxon signed-rank post-hoc tests and conclude that the differences between \MethodName and the other methods were found to be statistically significant with $p$ \textless 0.01.

\subsubsection{Discussion: average ranking model results.}
Additional analysis of the results presented in the previous section yielded several interesting insights about the merits of \MethodName.

\textbf{\MethodName is more robust than the CaD and distance-based methods}. Table \ref{tab:top3correlation} presents the five datasets for which \MethodName and the other state-of-the-art methods have the greatest differences on the SRC measure. In each case, \MethodName  outperforms CaD and distance-based by a significant margin. These results lead us to conclude that our method is capable of performing well in scenarios where the dataset in question has uncommon characteristics that may cause the other approaches to be ineffective (e.g., a small number of instances or missing values). In cases such as these, the embedding representations obtained from the transformation of the datasets to graphs in our method provide better performance for the average ranking model than the predetermined meta-features used by the other leading approaches examined.

\begin{table}[h!]
    \centering
    \begin{adjustbox}{max width=\textwidth}
    \begin{tabular}{l l l l}
    \hline
         Method Compared to \MethodName&Dataset Name& Improvement in SRC & Suggested Reason  \\
         \hline
        CaD & jobs& 0.303& Combination of nominal and numeric features + small number of instances\\
        
        CaD&autoUniv-au7-500& 0.278& Combination of nominal and numeric features\\
         CaD&analcatdata\_reviewer&0.257& Missing values\\

         Distance-based&analcatdata\_reviewer&0.349& Missing values \\
       Distance-based &autoUniv-au7-500& 0.278& Combination of nominal and numeric features\\

         \hline
    \end{tabular}
    \end{adjustbox}
    \caption{Average ranking model results - the five datasets with the greatest improvement in the SRC measure between \MethodName, CaD, and distance-based. Positive values indicate better performance by \MethodName.}
    \label{tab:top3correlation}
\end{table}

% \begin{table}[h!]
%     \centering
%     \begin{adjustbox}{max width=\textwidth}
%     \begin{tabular}{l c l}
%     \hline
%          Dataset Name& Improvement in SRC & Suggested Reason  \\
%          \hline
%           {\color{red} telecom\_churn}& {\color{red}0.226}& {\color{red}Large number of instances + combination of nominal and numeric features}\\
%          {\color{red}audit}&{\color{red}0.212}& {\color{red}Large number of features}\\
%          {\color{red}hayes-roth}& {\color{red}0.201}& {\color{red}Small number of instances and features}
%   \\
%          \hline
%     \end{tabular}
%     \end{adjustbox}
%     \caption{Average ranking model results - the top-3 datasets with the greatest improvement in the SRC measure between \MethodName and the distance-based method. Positive values indicate better performance by \MethodName.}
%     \label{tab:top3distance}
% \end{table}

\textbf{\MethodName is consistently better at identifying the optimal algorithm.} In addition to achieving the top MRR score {(0.822)}, our approach recommended the best algorithm (the algorithm in the first rank position) on 155 of the 210 datasets evaluated. As shown in Table \ref{tab:optimal_algorithm}, this value is significantly higher than that of the other methods. 

\begin{table}[h!]
    \centering
    \begin{tabular}{l c}
    \hline
         Method Name& Number of Datasets for which the Best Algorithm was Recommended \\
         \hline
         distance-based & {64} {(30.4\%)}  \\
         CaD & {65}   { (30.9\%)}\\
         \MethodName & \textbf{{155} {(73.8\%)}}\\
        \hline
    \end{tabular}
    \caption{Average ranking model results - the number of datasets for each method for which the optimal algorithm is recommended.  The best results are  highlighted.}
    \label{tab:optimal_algorithm}
\end{table}

\begin{figure}[h!]
    \centering
     \begin{adjustbox}{max width=0.6\textwidth}
    \includegraphics{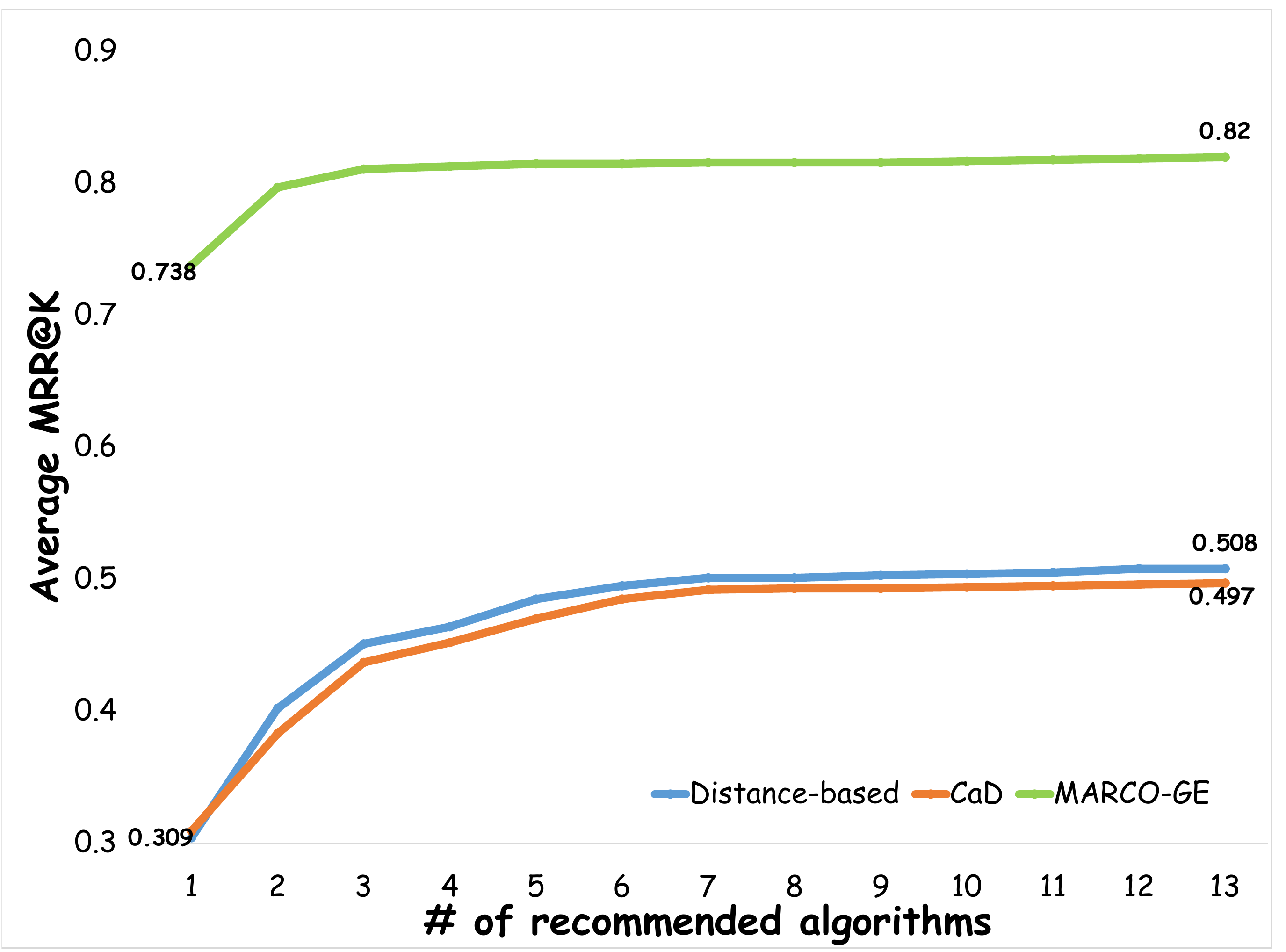}
    \end{adjustbox}
    \caption{Average ranking model results - average MRR@K over 210 datasets obtained by distance-based, CaD, and \MethodName vs the number of recommended algorithms.}
    \label{fig:MRR@k_averagemodel}
\end{figure}
\textbf{\MethodName is consistently better than the distance-based and CaD methods.} In order to determine whether the overall recommendation quality of \MethodName is better than the distance-based and CaD methods, we analyze these approaches based on their top-X ranked algorithms. In Figure \ref{fig:MRR@k_averagemodel}, we present the MRR@K results when choosing the best-performing algorithm of the top-K recommendations. While all of the methods show improved performance, \MethodName achieves an MRR@K greater than 0.8 after evaluating only three algorithms. In addition, it can be seen that our method converges after evaluating four algorithms, whereas the other methods converge after evaluating six algorithms. From the end-user perspective, \MethodName has beneficial added value: to find an optimal algorithm for a given problem, fewer algorithms need to be considered. 

{
\subsubsection{Analysis of the number of clusters}
In this section, the distribution of the number of clusters (K) hyperparameter is analyzed.
As discussed in subsection \ref{hyperparameter_tuning}, the value of $K$ is optimized for every dataset and clustering algorithm combination. 
The following algorithms were recommended by \MethodName as the most appropriate algorithms for the various datasets in the average ranking model: EAC, SL, KM, AL, CL, and WL. The distribution, median, and range of the values of $K$ for each algorithm are represented in a box plot in Figure \ref{fig:box_plot_number_of_clusters}. In addition, we divided the range of $K$ into four sub-ranges and presented the number of times an algorithm is recommended for each interval in Figure \ref{fig:hist_number_of_clusters}. 
 
From Figures \ref{fig:box_plot_number_of_clusters} and \ref{fig:hist_number_of_clusters}, we can conclude that: (1) the median value of $K$ over all of the algorithms is less than 10; (2) the interquartile ranges of algorithms SL and CL are smaller than the other algorithms; thus, these two algorithms do not usually require many search iterations to find the best number of clusters; (3) as can be seen from Figure \ref{fig:hist_number_of_clusters}, AL is the algorithm most recommended by \MethodName, followed by KM, SL, EAC, CL, and WL;
(4) algorithms SL, KM, AL, CL, and WL tend to provide better clustering solutions when $K$ is less than five; and 
(5) although AL is most successful when $K < 5$, it still has good performance for all of the other sub-ranges. 
\begin{figure}[h!]
    \centering
     \begin{adjustbox}{max width=0.6\textwidth}
    \includegraphics{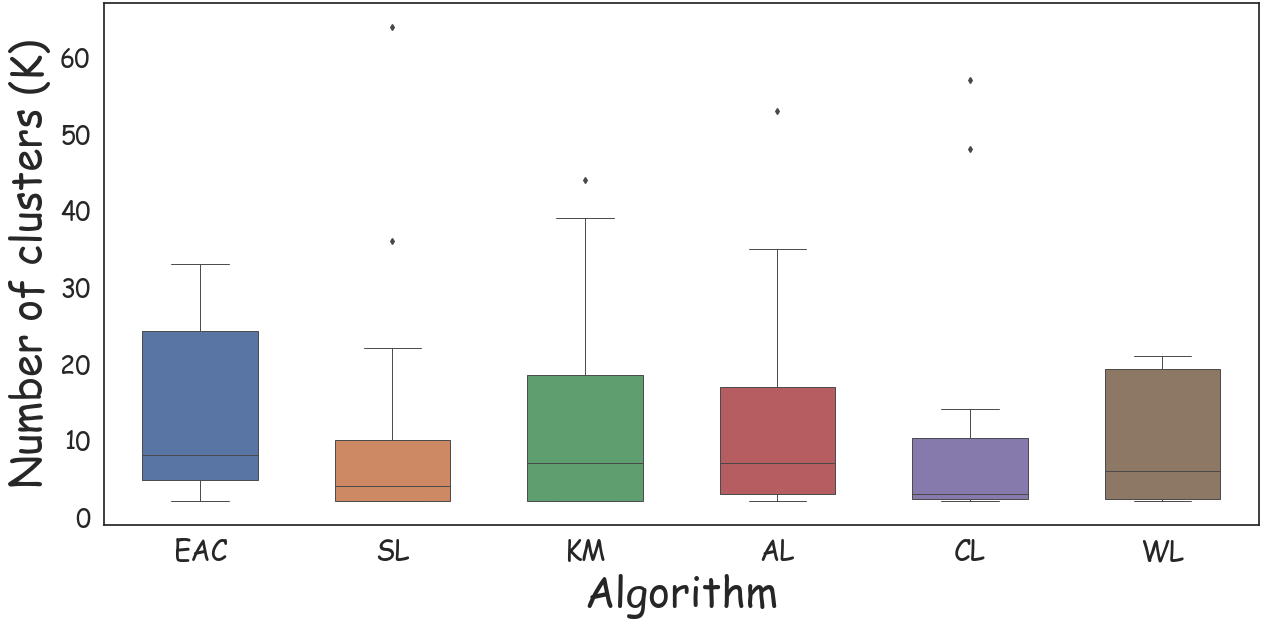}
    \end{adjustbox}
    \caption{{Average ranking model results - box plot depicting the distribution of K over the best recommended algorithms. For each box plot, the middle line represents the median, the lower bound indicates the first quartile, the upper bound represents the third quartile, and the whiskers' length is 1.5 times the interquartile range.}}
    \label{fig:box_plot_number_of_clusters}
\end{figure}

\begin{figure}[h!]
    \centering
     \begin{adjustbox}{max width=0.6\textwidth}
    \includegraphics{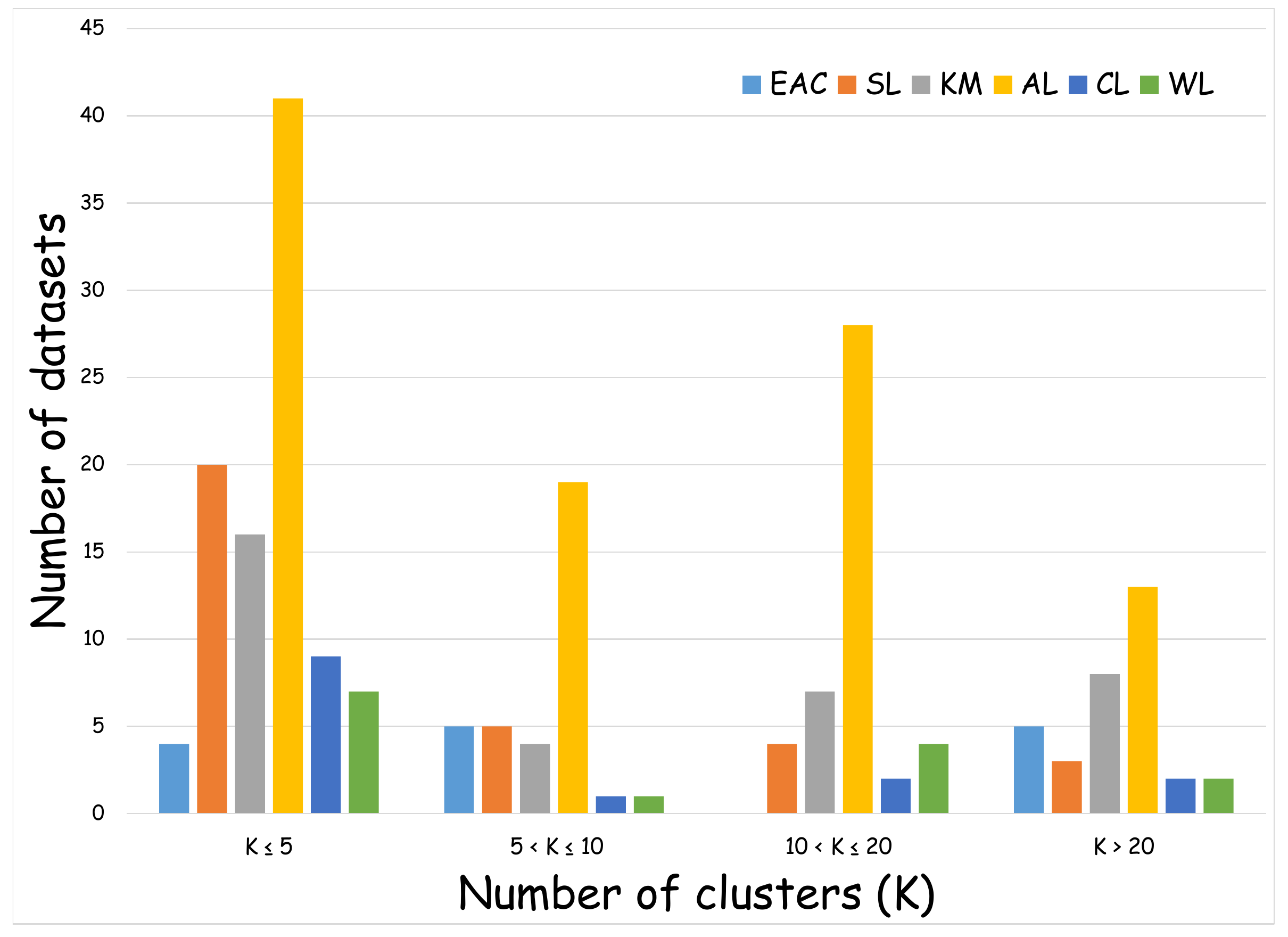}
    \end{adjustbox}
    \caption{{Average ranking model results - histogram presenting the distribution of $K$ for the best recommended algorithms across 210 datasets.}}
    \label{fig:hist_number_of_clusters}
\end{figure}

}

%% file: Method/clustering_algorithms.tex
\subsubsection{Clustering algorithms}\label{clustering_algorithm}
In this paper,  {17} widely used clustering algorithms are evaluated on a diverse number of datasets.\footnote{ The implementation of several algorithms and several  internal clustering measures are taken from \newline https://github.com/ItayGabbay/ClusteringAlgorithmSelection.} These algorithms enable comprehensive  analysis from different perspectives. The algorithms selected represent various clustering algorithm categories, as seen in Table \ref{tab:categories}.
The clustering algorithms evaluated are:

\textbf{1) Evolutionary Algorithm for Clustering} (EAC) - a genetic algorithm which is capable of automatically finding the optimal number of clusters for a given dataset \cite{alves2006towards}. The solution quality (fitness) is evaluated by using the silhouette metric.

\textbf{2) Particle Swarm Optimization for Clustering} (PSC) - a biologically-inspired algorithm motivated by an analogy to a flock of birds \cite{van2003data}. This algorithm maintains a set of potential solutions, i.e., particles, where each particle represents a solution to an optimization problem. In clustering a particle represents a potential cluster's center. The particles move in the search space and are ultimately positioned and divided according to the natural groups of data. The number of clusters is an input parameter.

\textbf{3) Minimum Spanning Tree} (MST) - used to  detect clusters with a heterogeneous nature \cite{4031882}. Given a dataset, the algorithm generates a complete graph $G$ of the dataset. Then, an MST is constructed from $G$. By removing the edges that satisfy a predefined inconsistency criterion, a solution with $k$ clusters is produced.

\textbf{Hierarchical Agglomerative Clustering} - a bottom-up approach that generates a series of models with a number of clusters from $n$ (each data point is an individual cluster) to one (all of the data points in one cluster) \cite{murtagh2012algorithms}. In each step, the approach merges the pair of clusters that minimally increases a given linkage distance. The linkage criterion determines which distance should be used between data point sets. Methods 4, 5, 6, and 7 are examples of hierarchical agglomerative algorithms with different linkage criteria.

\textbf{4) Agglomerative Single-Linkage} (SL) - merges the two clusters with the minimal pairwise distance \cite{xu2005survey}. 

\textbf{5) Agglomerative Average-Linkage} (AL) - combines the clusters  $C_i$ and $C_j$ with the minimal average distance between a pair of points, where one point is from cluster $C_i$, and the other point is from cluster $C_j$.  

\textbf{6)  Agglomerative Complete-Linkage} (CL) links the two clusters with the shortest distance between the points farthest from each cluster, i.e.,  with the smallest diameter \cite{defays1977efficient}. 

\textbf{7) Agglomerative Ward-Linkage} (WL) - finds the pair of clusters that lead to the minimum increase in the variance after merging.

\textbf{8) K-Means} (KM) - the most commonly used simple clustering method \cite{jain2010data}. This algorithm clusters data points into a predefined number of groups based on the similarities between them. This technique is sensitive to the initialization of the clusters' centers.

\textbf{9) K-Harmonic Means} (KHM) - a technique that overcomes the K-means method's problem regarding initialization sensitivity. Using the harmonic mean, this algorithm defines the clusters' centers \cite{zhang2000k}; thus, the solution provided is independent of the initialization of the clusters' centers. The number of clusters is an input parameter. 

\textbf{10) Kernel K-Means} (KKM) - an algorithm that follows the same procedure as the KM, except a different distance calculation method is used \cite{dhillon2004kernel}. In this case, a kernel method is used instead of the Euclidean distance to find clusters that are nonlinearly separable in the input space. 

\textbf{11) Mini Batch K-means} (MBK) - a variant of the KM algorithm that utilizes mini-batches to improve the algorithm's efficiency in terms of computation time. 

\textbf{12) Fuzzy C-Means} (FC) - allows each data point to belong to multiple clusters by assigning a membership score to each data point that corresponds to each cluster's center. The membership score $M_{i, C_j}$ is determined on the basis of the distance from point $i$ to cluster center $C_j$.
 The closer the data point is to the cluster's center, the greater its membership score for the particular cluster center  \cite{man1994detection}.

\textbf{13) Density-Based Spatial Clustering Applications with Noise} (DBSCAN) - an algorithm that effectively discovers different shaped clusters; it also deals with noise and outliers \cite{ester1996density}. The solution is produced by combining dense areas separated by regions of sparse areas. A distance threshold $Eps$ is an input parameter. By using a heuristic that defines a small value for $Eps$ and increasing it by 10\% in each iteration, we achieve a partition where the outliers are less than 10\% of the data.

{ \textbf{14) Mean Shift} (MS) - a centroid-based algorithm that aims to find high-density regions in the data \cite{comaniciu2002mean}. MS involves iteratively updating the centroid candidates so that they equal the mean of the points within a given region. Finally, MS filters the candidate set so that near-duplicate centroids are removed. }  

{ \textbf{Gaussian Mixture Models} - probabilistic models that assume that the entire dataset is modeled by a mixture of a finite number of Gaussian distribution components, and each of these components represents a cluster \cite{reynolds2009gaussian}. Algorithms 15, 16, and 17 are variants of Gaussian mixture models that use different types of covariance matrices. }

{ \textbf{15) Gaussian Mixture Models with a Full Matrix} (GMF) - use  a covariance matrix for each component. }

{ \textbf{16) Gaussian Mixture Models with a Tied Matrix} (GMT) - share the same covariance matrix among the Gaussian components.}

{ \textbf{17) Gaussian Mixture Models with a Diagonal Matrix} (GMD) - use a diagonal covariance matrix for each component.}

 \begin{table}[h!]
    \centering
    \begin{tabular}{l l}
    \hline
         \textbf{Category} &\textbf{Algorithms} \\
         \hline
         Bio-inspired& EAC, PSC\\
         Graph-based& MST\\
         Hierarchical &SL, AL, CL, WL\\
         Partitional&KM, KHM, KKM, MBK\\
         Fuzzy & FC \\
         Density-based& DBSCAN\\
        {Expectation-maximization clustering} & {GMF, GMT, GMD} \\
        {Kernel density estimation} & {MS}\\ 
         \hline
         
    \end{tabular}
    \caption{The clustering algorithm categories evaluated. }
    \label{tab:categories}
\end{table}
{

\subsubsubsection{Hyperparameter tuning}\label{hyperparameter_tuning}
\setlength{\parindent}{1em}
The performance of the abovementioned algorithms depends on the optimization of their hyperparameters.
In our experiments, we optimize the hyperparameters of the clustering algorithms using Hyperopt \cite{bergstra2013hyperopt} (with the Tree-structured Parzen Estimator (TPE) algorithm). Hyperopt is a Python library that searches for the optimal hyperparameter configuration based on Bayesian optimization.  It intelligently investigates the hyperparameter search space so that the next configuration to be executed is selected based on the performance of the previous trials. Hyperopt requires an \textit{objective function} to minimize and a \textit{search space} for each hyperparameter.

The first element is the objective function to minimize; this is used to assign loss scores to the configurations tested.  Since we assume the ground truth labels are not known, the objective function should rely on internal measures. Although several internal measures are proposed (e.g., silhouette and Dunn indices), each of them evaluates different aspects of the clustering solution, and therefore none of them is applicable for all cases \cite{poulakis2020autoclust, van2015using}. Furthermore, to the best of our knowledge, there is no method that combines the values of multiple internal measures into one score. Generating a score that quantifies all of the internal measures is not straightforward due to the fact that in most cases, the measure's range is not normalized.   

To address this challenge, we present a procedure that maps the evaluation values produced by multiple internal measures into a single score. The procedure, which is described in Algorithms \ref{Alg:Total} and \ref{Alg:Individual} and Figure \ref{fig:flowcharts}, applies three steps for each configuration tested $c_j$, where $j$ = (0,..., number of trials - predetermined by the user):
\begin{enumerate}
  
    \item  By using all of the internal measures presented in the subsection that follows (we denote the set of internal measures as $M$), we evaluate the clustering solution produced by $c_j$. As a result, $evaluation_{m,j}$ is produced for each $m\in M$.
    \item  We normalize $evaluation_{m,j}$ using min-max scaling, where the minimum and maximum values are based on the evaluation values obtained by $m$ so far.  We denote the normalized value of each measure as $score_{m,j}$.
    \item  The normalized values of each $m\in M$ are merged by averaging the values into a single score. Since Hyperopt minimizes the objective function, we multiply the average score by -1, and the calculation of the $trial\_score_j$ is completed.
\end{enumerate}

It is important to note the following four points regarding the procedure described above. First, $M$ includes internal measures with  objective functions to minimize; thus, the evaluation result $evaluation_{m,j}$ of these measures is multiplied by -1. Second, the $score_{m,0}$ of each $m \in M$ is initialized with 0.5; thus,  $trial\_score_0 = 0.5 $. Third, given a trial $j$, we update the value of the corresponding $score_{m,j}$ for each $m \in M$, where $j = (0,...,j-1)$, based on the new scaling. Finally, for large datasets, we randomly sampled 1000 instances as part of the hyperparameter tuning.

\input{Algorithms/Total_score}
\input{Algorithms/Measure_score}

% \begin{figure}
%      \centering
%      \begin{subfigure}[b]{0.4\textwidth}
%          \centering
%          \includegraphics[width=\textwidth]{Figures/main.pdf}
%          \caption{Flow chart of Algorithm 3}
%          \label{fig:y equals x}
%      \end{subfigure}
%      \hfill
%      \begin{subfigure}[b]{0.4\textwidth}
%          \centering
%          \includegraphics[width=\textwidth]{Figures/sub.pdf}
%          \caption{Flow chart of Algorithm 4}
%          \label{fig:three sin x}
%      \end{subfigure}
    %  \hfill
    %  \begin{subfigure}[b]{0.3\textwidth}
    %      \centering
    %      \includegraphics[width=\textwidth]{graph3}
    %      \caption{$y=5/x$}
    %      \label{fig:five over x}
    %  \end{subfigure}
    %     \caption{Three simple graphs}
    %     \label{fig:three graphs}
% \end{figure}
% \includepdf[pages={1},picturecommand*={\put (\LenToUnit{.05\paperwidth},20) {Page 3 in the original.};}]]{Figures/flowchart.pdf}

% \begin{figure}[]
%     \centering
%     \includegraphics[width=1\textwidth]{Figures/flowchart.pdf}
%     \caption{The flow charts of the algorithms}
%     \label{fig:my_label}
% \end{figure}

The second element is the search space for each hyperparameter. Based on our experience with clustering problems, we first select several significant hyperparameters for each algorithm and then specify a plausible search space for each of them. Table \ref{tab:parameters} summarizes the hyperparameters examined for each algorithm. As can be observed in Table \ref{tab:parameters}, in most cases, the number of clusters (K) hyperparameter serves as the input for the clustering algorithm.  Determining the right number of clusters has a critical effect on the clustering solution. Consequently, we focus the search space of K in the following way. 
First, we perform hyperparameter optimization (using Hyperopt) for the four algorithms that do not require the value of $K$ in advance (i.e., EAC, MST, DBSCAN, and MS).   Then, for each of these algorithms,  we  select the clustering solution of the best configuration found and its evaluation score.  Next, the search  space of  $K$ is specified within the range of the number of clusters of the solution that achieved the  best evaluation score.

 \begin{table}[h!]
    \centering
  \begin{adjustbox}{max width=\textwidth}
    {\begin{tabular}{l l}
    \hline
         \textbf{Algorithm} &\textbf{Hyperparameters} \\
         \hline
        EAC& Initial number of clusters; population size \\
        
         PSC& Number of particles; maximum velocity \\
         MST & Cut off scale; minimum cluster size\\
         SL, AL, CL, WL& Number of clusters\\
         KM, KHM, KM, KHM, KKM, MBK& Number of clusters; maximum number of iterations\\
         FC & Number of clusters; maximum number of iterations; m - the fuzziness parameter \\
        DBSCAN& Eps (neighborhood radius); minimum number of samples\\
        MS & Bandwidth\\
        GMF, GMT, GMD & Number of clusters; maximum iterations \\
         \hline
         
    \end{tabular}}
    \end{adjustbox}
    \caption{{Hyperparameters of the clustering algorithms evaluated.}}
    \label{tab:parameters}
\end{table}

\newpage
\begin{figure}[h!]
 \caption{The flow charts of the algorithms}
\includepdf[pages={1}]{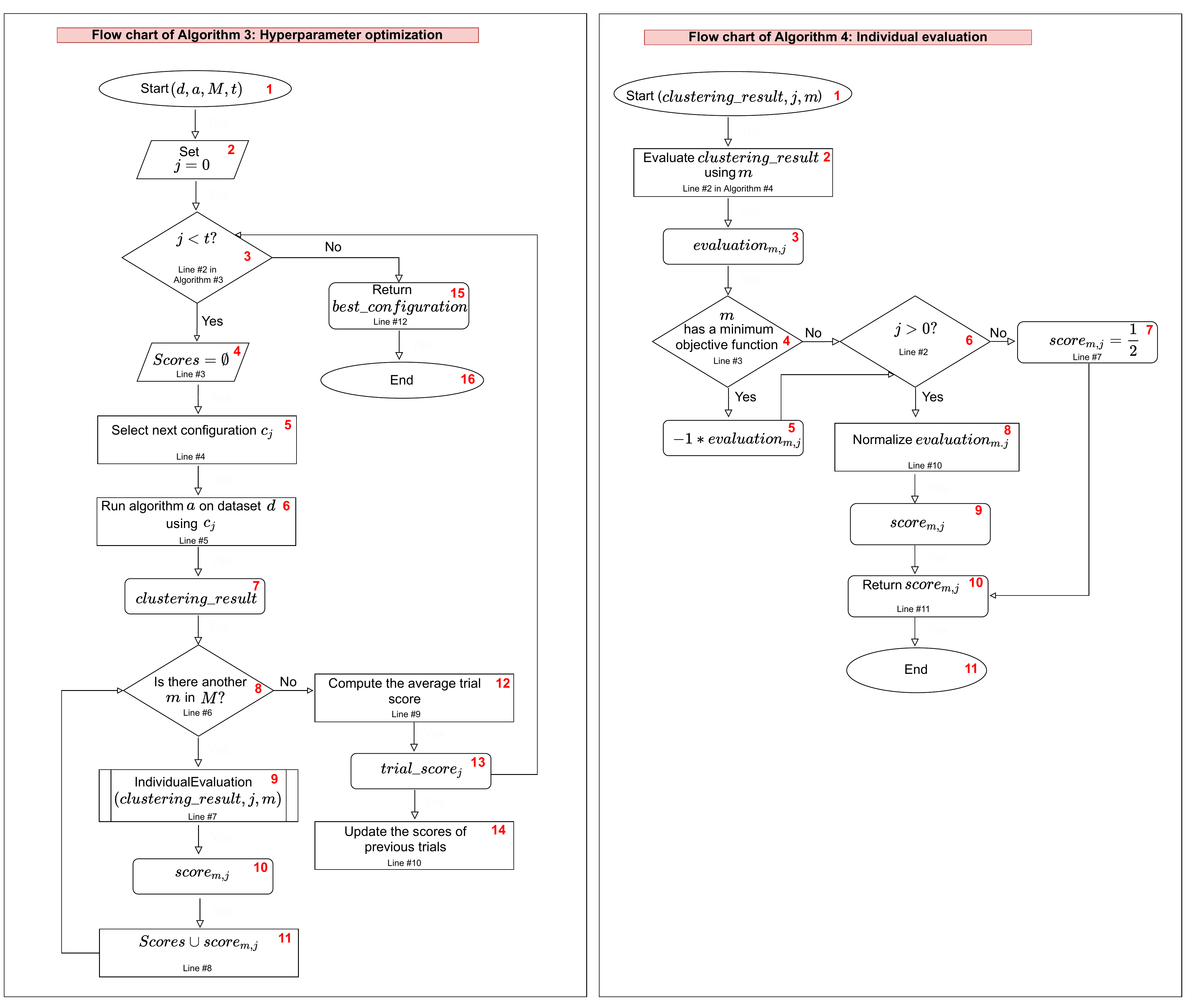}
    \subfloat[\label{3}Flow chart of Algorithm 3]{\hspace{.4\linewidth}}
    \hspace{3cm}
    \subfloat[\label{4} Flow chart of Algorithm 4]{\hspace{.4\linewidth}}
    \label{fig:flowcharts}
\end{figure}
\newpage

\begin{figure}[h!]
    \centering
    \begin{adjustbox}{max width=0.9\textwidth}
    \includegraphics{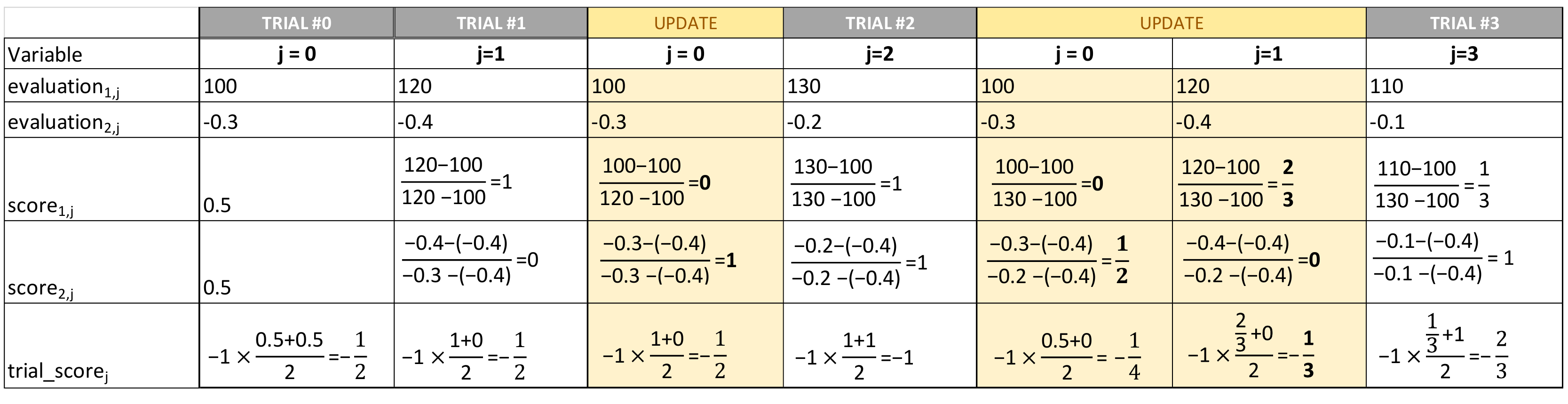}
    \end{adjustbox}
    \caption{{The process of updating the evaluation score as part of the hyperparameter tuning.}}
    \label{fig:hyperparmeter_opt}
\end{figure}

Figure \ref{fig:hyperparmeter_opt} illustrates some trials of the process of updating the evaluation score as part of the hyperparameter optimization. Consider a case in which an algorithm $a$ is tested on dataset $d$ with configuration $c_0$ (box 6, Figure \ref{3}). This test produced the solution $clustering\_result$ (box 7, Figure \ref{3}). Two internal measures, $m_1$ and $m_2$, were used to assess the $clustering\_result$. Assume that the goal of $m1$ is maximization, while $m_2$ aims to minimize.  
In trial \#0, where $j=0$, $m_1$ is utilized to evaluate the $clustering\_result$ (box 2, Figure \ref{4}), and the evaluation value obtained, which is denoted as $evaluation_{1,0}$, is 100 (box 3, Figure \ref{4}). Since this is the initial trial, $score_{1,0} = \frac{1}{2}$ (box 7, Figure \ref{4}), and this is provided to Algorithm 3 (box 10, Figure \ref{4}). Next, measure $m_2$ is applied to evaluate the $clustering\_result$ (box 2, Figure \ref{4}), and the variable $evaluation_{2,0}$ is assigned with the value of $0.3$ (box 3, Figure \ref{4}). $m_2$ has a minimum objective function; hence, the variable $evaluation_{2,0}$ is multiplied by -$1$ (box 5, Figure \ref{4}). Again, this is the initial trial; thus, $score_{2,0} = 0.5$ (box 7, Figure \ref{4}). $score_{2,0}$ is also provided to Algorithm 3 (box 10, Figure \ref{4}). Now that all of the available measures have been applied,  $trial\_score_0$ is computed by averaging $score_{1,0}$ and $score_{2,0}$ and multiplying the result by -$1$, i.e., -$1 \times \frac{0.5 + 0.5}{2} =$ -$0.5$ (box 12, Figure \ref{3}).

In the next trial, another configuration is tested, and the variable $clustering\_result$ is assigned with a new value (box 7, Figure \ref{3}). Based on the assessment of $m_1$ on the $clustering\_result$ (box 2, Figure \ref{4}), the value of $evaluation_{1,1} $ is determined to be 120. Since this is the second iteration, i.e., $j=1$, we normalized the value of $evaluation_{1,1}$ using min-max scaling (box 8, Figure \ref{4}). The minimum and the maximum values obtained by $m_1$ so far are 100 and 120, respectively. Consequently, the normalized value, $score_{1,1}$, is equal to $\frac{120 \mbox{-} 100}{120\mbox{-}100} = 1$ (box 9, Figure \ref{4}). Afterwards, the evaluation value obtained by $m_2$ is 0.4, i.e., $evaluation_{2,1} = 0.4$ (box 3, Figure \ref{4}). The value of $evaluation_{2,1}$ is multiplied by -1 (box 5, Figure \ref{4}); thus, $evaluation_{2,1} =$ -$0.4$. Here, the minimum and maximum values obtained by $m_2$ are -$0.4$ and -$0.3$, respectively. Hence, the normalized score $score_{2,1}$ is equal to $\frac{\mbox{-}0.4\mbox{-}(\mbox{-}0.4)}{\mbox{-}0.3\mbox{-}(\mbox{-}0.4)} =0 $ (box 9, Figure \ref{4}). Notably, measure $m_1$ achieved better results in trial \#1 than trial \#0; thus, the value of $score_{1,1}$ increased. In contrast, in trial \#1 measure $m2$ obtained worse results than the initial trial; therefore, the value of $score_{2,1}$ decreased. Since there are no other measures that have not been tested, the $trial\_score_1$ is computed (box 12, Figure \ref{3}). As a result, $trial\_score_1 = \mbox{-}1 \times \frac{1+0}{2} = \mbox{-}0.5$ (box 13, Figure \ref{3}).

At this point, for each $m\in M$, where $j=(0,...,j-1)$, i.e, $j=(0)$, the value of the corresponding $score_{m,j}$ is updated based on the new scaling (box 14, Figure \ref{3}). In this case, the minimum and maximum values obtained by $m_1$ are 100 and 120, respectively, where the minimum and maximum values obtained by $m_2$ are -0.4 and -0.3, respectively. Therefore, $score_{1,0} = \frac{100 \mbox{-} 100}{120\mbox{-}100} = 0$, and $score_{2,0} = \frac{\mbox{-}0.3\mbox{-}(\mbox{-}0.4)}{\mbox{-}0.3\mbox{-}(\mbox{-}0.4)} = 1$. Based on the updated values, we recomputed the value of $trial\_score_0$, which is equal to $ \mbox{-}1 \times \frac{1+0}{2} = \mbox{-}0.5$.
 This process continues until the number of trials ($t$) has been completed (box 16, Figure \ref{3}).  
}

%% file: Algorithms/Total_score.tex
\setcounter{algocf}{2}
\begin{algorithm}[h!]
\SetAlgoLined
\DontPrintSemicolon
\KwInput{$d \in D$ - a dataset, $a$ -an algorithm , $M$ - a set of internal measures, $t$ - number of trials}
  \KwOutput{$best\_configuration$ - The best configuration }

    \SetKwFunction{FMain}{HyperparameterOptimization}
    \SetKwProg{Fn}{Function}{:}{}
    \Fn{\FMain{$d$, $a$, $M$, $t$}}{
    
     \For{$j\gets0$ \KwTo $t$ }{
     $ Scores   \longleftarrow \emptyset$ \\
     $ c_j   \longleftarrow Hyperopt.getNextConfiguration()$ \\
      $ Clustering\_result  \longleftarrow a(c_j, d)$ \\
    
     \ForEach{$ m \in M $}
    { 
     $score_{m,j}  \longleftarrow IndividualEvaluation(Clustering\_result, j, m)$ \\ 
     $Scores \longleftarrow Scores \cup score_{m,j}  $\\

    }
    $trial\_score_j \longleftarrow Scores.mean() \times -1 $\\
    $Hyperopt.trials.UpdateScores(trial\_score_j)$

    }
    $best\_configuration \longleftarrow Hyperopt.GetBestConfiguration()$\\
\textbf{return} $best\_configuration$ 
}
\textbf{End Function}
\caption{ Hyperparameter optimization }
\label{Alg:Total}
\end{algorithm}
\normalsize

%% file: Algorithms/Measure_score.tex
\setcounter{algocf}{3}
\begin{algorithm}[h!]
\SetAlgoLined
\DontPrintSemicolon
\KwInput{$Clustering\_result$ , $j$ - the index of the current trial, $m$ - an evaluation measure}
  \KwOutput{$score_{m,j}$ - a normalized individual evaluation score of internal measure $m$ }

   \SetKwFunction{FMain}{IndividualEvaluation}
    \SetKwProg{Fn}{Function}{:}{}
    \Fn{\FMain{$Clustering\_result$, $j$, $m$}}{
    $evaluation_{m,j} \longleftarrow m.getEvaluation(Clustering\_result)$\\
     \uIf{  $m$ has a minimum objective function}{
    $evaluation_{m,j} \longleftarrow evaluation_{m,j} \times -1 $
    }
    % $max\_prev\_evaluation \longleftarrow m\_i.getPreviousEvaluations().max()$\\
    % $min\_prev\_evaluation \longleftarrow m\_i.getPreviousEvaluations().max()$\\
    $m.appendEvaluation(evaluation_{m,j})$\\
    
    \eIf{j $=$ 0}
{ 
   $ score_{m,j} = \frac{1}{2}$
}
{
    $evaluations  \longleftarrow m.getAllEvaluations()$ \\
   
     $ score_{m,j} \longleftarrow \frac{evaluation_{m,j} - evaluations.min() }{evaluations.max() - evaluations.min()}$\\
    %  $Prev\_individual\_scores \longleftarrow trials.getPreviousScores(m_i)$\\

    % \uIf{$evaluations[t-1]  \neq Current\_evaluation$}{

%   }
   
%   \uIf{  Current\_evaluation $<$  min\_prev\_evaluation}{
% \For{$b\gets0$ \KwTo $j-1$ }{
% $Prev\_individual\_scores[b] \longleftarrow     \frac{evaluations[b] -  Current\_evaluation }{evaluations.max() - Current\_evaluation}$   \\
% }
% $trials.update(m_i, Prev\_individual\_scores)$\\
% }

% \uElseIf{Current\_evaluation $>$  max\_prev\_evaluation}
% {
% \For{$b\gets0$ \KwTo $j-1$ }{
% $Prev\_individual\_scores[b] \longleftarrow     \frac{evaluations[b] -  evaluations.min() }{Current\_evaluation - evaluations.min()}$   \\
% }
% $trials.update(m_i, Prev\_individual\_scores)$
% }
}

%  $m_i.appendScore(Current\_individual\_score)$\\
\textbf{return} $score_{m,j}$ 
}
\textbf{End Function}
\caption{Individual evaluation}
\label{Alg:Individual}
\end{algorithm}

%% file: Method/Internal_indices.tex
\subsubsection{Internal clustering evaluation measures}\label{clustering_evaluation_measure}
 Since clustering is an unsupervised task, most of the relevant problems do not have a priori knowledge about the solution of a given clustering problem. Thus, in this study, we chose 10 widely used internal measures that represent different aspects such as cluster compactness and cluster separation. 
 
As mentioned above (subsection \ref{clustering_performance_assessment}), some internal measures follow the notions of inter/intra-cluster measures. While intra-cluster measures
assess cluster compactness, inter-cluster measures assess the quality of the separation between clusters.
 There are several definitions of inter/intra measures available. In this work,  we chose the largest diameter of a cluster as the intra-cluster measure and the sum of pairwise distances between data points from two different clusters as the inter-cluster measure. All of the indices use the $Euclidean$ distance between data points. 

The indices are:

1) \textbf{Davies-Bouldin} \cite{liu2010understanding} is a function of the average similarity between each cluster and the cluster most similar to it.  In the context of this index, the similarity is defined as a measure that balances intra-cluster scatter against inter-cluster separation. The better the clusters, the less the value of the index.

2) \textbf{Calinski-Harabasz} \cite{liu2010understanding}, also known as the \textit{variance ratio criterion}, is defined as the ratio of between-cluster dispersion and inter-cluster dispersion for all clusters. The higher the score, the better the clusters' quality.  

3) \textbf{Bezdek-Pal} \cite{bezdek1998some} is an average value of all of the inter-cluster distances. The better the partition, the higher the value of the index.

4) \textbf{Dunn index} \cite{dunn1973fuzzy} computes the ratio between the smallest separation value and the largest compactness value. Maximizing this ratio leads to a better solution.

5) \textbf{Silhouette score} \cite{liu2010understanding} is a measure of how similar a data point is to the cluster it belongs to, compared to its similarity to other clusters. The best value is 1 and the worst value is -1. A high value implies that the data point is well matched to its own cluster, whereas a value close to -1 indicates that the point is assigned to the wrong cluster.

6) \textbf{Milligan-Cooper} \cite{milligan1985examination} is a maximization criterion based on the point biserial index, which is a statistical correlation measure between continuous and binary variables.  For each pair of points, a point biserial correlation is calculated between the corresponding values in the distance matrix and a dummy variable which indicates whether or not they were assigned to the same cluster.  A value of zero is assigned if the two corresponding points are clustered together by the algorithm, otherwise, a value of one is assigned.

7) \textbf{Handl-Knowles-Kell} \cite{handl2005computational} assesses how well a given clustering solution follows the idea of grouping neighboring data points together in the same cluster.  The value of the index increases when there are  differences in the clustering of some data points and the clustering of their neighbors.

8) \textbf{Hubert-Levin} \cite{hubert1976general} evaluates whether similar data points share the same cluster and whether data points with a large distance between them are assigned to different clusters. Clustering quality increases when the value of the index decreases.

9) \textbf{SD-Scat}  \cite{halkidi2001clustering} measures the average compactness of clusters by computing the scattering within clusters. A small value indicates compact clusters.

10) \textbf{Xie-Beni} \cite{liu2010understanding} is defined as the ratio of the overall variation to the smallest distance between the clusters. When the data has been appropriately clustered, the index value should be low.

Table \ref{tab:internal_indices} presents the internal indices, their respective domains, objectives, and mathematical formulations. We use those algorithms and indices to create meta-knowledge.

\begin{table}[h]
    \centering
    \begin{adjustbox}{max width=\textwidth}
    \begin{tabular}{l c l l}
    \hline
        \Huge{\textbf{Index}} & \Huge{\textbf{Score Domain}} & \Huge{\textbf{Objective}} &\Huge{\textbf{Mathematical Formulation} }\\
        \hline
        \Huge{Bezdek-Pal}& \Huge{$[0;\infty]$} & \Huge{Max} &\Huge{$\sum_{i=1}^{NC}\sum_{j=i+1}^{NC}\frac{\sum_{x\in C_i, y \in C_j}d(x,y)}{|n_i||n_j|}$} \\
         \Huge{Dunn index} & \Huge{$[0,\infty]$} & \Huge{Max} & \Huge{$\min_i \textbf{\{} \min_j(\frac{\min_{x\in C_i, y \in C_j}d(x, y)}{\max_k \{ \max_{x, y\in C_k}d(x,y)\}})\textbf{\}}$}\\
         \Huge{Calinski-Harabasz} & \Huge{$[0,\infty]$} & \Huge{Max} & \Huge{$\frac{\sum_i n_i d^2(c_i, c)/ (NC-1)}{\sum_i\sum_{x\in C_i} d^2(x, c_i) / (n-NC)}$}\\
         \Huge{Silhouette score} & \Huge{$[-1,1]$} &\Huge{ Max}&\Huge{$ \frac{1}{NC} \sum_i\{ \frac{1}{n_i}\sum_{x \in C_i} \frac{b(x) - a(x)}{\max[b(x),a(x)]}\}$}\\
         \Huge{Milligan-Cooper} & \Huge{$[-1,1]$} & \Huge{Max} & \Huge{$\frac{(\frac{\sum_{i=1}^{NC}\sum_{j=i+1}^{NC}\sum_{x\in C_i, y \in C_j}d(x,y)}{n_b} - \frac{\sum_{i=1}^{NC} \sum_{x,y \in C_i, l \textless m} d(x_l,y_m)}{n_w})\sqrt{\frac{n_w n_b}{{n_t}^2}}}{s_d}$}\\
        \Huge{Davies-Bouldin} & \Huge{$[0,\infty]$} &\Huge {Min} & 
             \Huge{$\frac{1}{NC}\sum_i \max_{j, j \neq i} \textbf{\{} \frac{\frac{1}{n_i} \sum_{x \in C_i} d(x, c_i) +\frac{1}{n_j} \sum_{x \in C_j} d(x, c_j)}{d(c_i, c_j)} \textbf{\}}$}\\
         \Huge{Handl-Knowles-Kell} & \Huge{$[0,\infty]$} &\Huge{Min} & \Huge{$ \sum_{i=1}^n \sum_{j=1}^{L} x_{i,nn_{i(j)}}, x_{i,nn_{i(j)}} = 0$ if i and j are in the same cluster and $\frac{1}{j}$ otherwise }\\
         \Huge{Hubert-Levin} &\Huge{ $[0;1]$} & \Huge{Min} &\Huge{$ \frac{\sum_{i=1}^{NC}\sum_{x,y \in C_i, l \textless m} d(x_l, y_m) -S_{min}}{S_{max} - S_{min}}, S_{min} \neq S_{max}$}\\ 
         \Huge{SD-Scat} & \Huge{$[0, \infty]$}& \Huge{Min} & \Huge{$\frac{1}{NC}\sum_i\frac{||\sigma(C_i)||}{||\sigma(D)||}$} \\
        \Huge{Xie-Beni} & \Huge{$[0, \infty]$} &\Huge {Min} & \Huge{$\frac{\sum_i \sum_{x\in C_i}d^2(x, c_i)}{n* \min_{i, j \neq i}d^2(c_i, c_j)}$} \\
        \hline
    \end{tabular}
    \end{adjustbox}
    \caption{Description of internal indices. Mathematical symbols: $D \mbox{-}$ dataset; $n\mbox{-}$ number of instances in $D$; $n_t\mbox{-}$ number of pairs of instances in $D$; $n_w\mbox{-}$ number of pairs of instances belonging to the same cluster; $n_b\mbox{-}$ number of pairs of instances belonging to different clusters; $c\mbox{-}$ center of $D$; $NC\mbox{-}$ number of clusters; $C_i\mbox{-}$ the i-th cluster; $n_i\mbox{-}$ number of instances in $C_i$; $c_i\mbox{-}$ center of $C_i$; $d(x,y)\mbox{-}$ distance between $x$ and $y$; $s_d\mbox{-}$ standard deviation of all distances; $\sigma(C_i)\mbox{-}$ variance vector of $C_i$; $L\mbox{-}$ parameter determining the number of neighbors to be considered; $S_{min}\mbox{-}$ sum of the $x$ smallest distances in $n_t$; $S_{max}\mbox{-}$ sum of the $x$ largest distances in $n_t$; $a(x)\mbox{-}$  average distance between the $x$th sample and all other samples included in the $i$th cluster; $b(x)\mbox{-}$  minimum average distance between the $x$th sample and all of the samples clustered in $k$th cluster for $k \neq i$ .}
    \label{tab:internal_indices}
\end{table}

%% file: Experiments/Experiments_individual_index.tex
\subsection{Evaluation results: individual model mode}
In the individual model mode evaluation, a unique model was created for each of the 10 internal measures. 
The mean values of the SRC and MRR measures for each index over 210 datasets are summarized in Tables \ref{tab:SRC_values_individual} and \ref{tab:MRR_values_individual}, respectively. 

\begin{table}[h!]
    \centering
    \begin{tabular}{ l c c c}
    \hline
        Index & Distance-Based & CaD & \MethodName  \\
        \hline
         Bezdek-Pal & 0.492& 0.500 &\textbf{0.532} \\
         Dunn Index & 0.544 & 0.531 &\textbf{0.568}\\
         Calinski-Harabasz& 0.638 & 0.635 &\textbf{0.669}\\
         Silhouette score & 0.487&0.487&\textbf{0.527}\\
         Milligan-Cooper & 0.428 & 0.410 & \textbf{0.456}\\
         Davies-Bouldin& 0.598& 0.602 & {\textbf{0.620}}\\
         Handl-Knowles-Kell&0.602 &0.586& {\textbf{0.624}}\\
         Hubert-Levin& 0.557&0.559& {\textbf{0.593}}\\
         SD-Scat& 0.488 & 0.484 & \textbf{0.524}\\
         Xie-Beni&0.502 &0.502&   \textbf{0.526}\\
         \hline
    \end{tabular}
   \caption{
  Individual model results - the average SRC  values of the evaluated methods over 210 clustering datasets. The best results with statistical significance are highlighted.}
    \label{tab:SRC_values_individual}
\end{table}

\begin{table}[h!]
    \centering
    \begin{tabular}{ l c c c}
    \hline
        Index & Distance-Based & CaD & \MethodName  \\
        \hline
         Bezdek-Pal & 0.628& 0.618 &\textbf{0.679} \\
         Dunn Index & 0.353 & 0.378 &\textbf{0.676}\\
         Calinski-Harabasz& 0.304 & 0.306 &\textbf{0.484}\\
         Silhouette score & 0.368&0.370&\textbf{0.529}\\
         Milligan-Cooper & 0.281 & 0.287 & \textbf{0.404}\\
         Davies-Bouldin& 0.651& 0.671 & \textbf{0.761}\\
         Handl-Knowles-Kell& 0.517 &0.524& \textbf{0.718}\\
         Hubert-Levin&0.320 &0.373& {\textbf{0.676}}\\
         SD-Scat& 0.440 & 0.470 &  \textbf{0.618}\\
         Xie-Beni& 0.482 & 0.471 &  \textbf{0.828}\\
         \hline
    \end{tabular}
   \caption{Individual model results - the average MRR  values of the evaluated methods over 210 clustering datasets. The best results with statistical significance are highlighted.}
    \label{tab:MRR_values_individual}
\end{table}

We used the Friedman test to validate the statistical significance of the differences between the results of the methods evaluated  \cite{demvsar2006statistical}; the results are presented in Tables \ref{tab:SRC_values_individual} and \ref{tab:MRR_values_individual}.  The null hypothesis that the three methods perform the same, and the observed differences are merely random was rejected with $p$ \textless 0.01 for almost all models (the { SRC of Xie-Beni and silhouette score were rejected with $p$ \textless 0.05}). We proceeded with Wilcoxon signed-rank post-hoc tests and conclude that the differences between \MethodName and all of the other methods were found to be statistically significant with $p$ \textless 0.01, {except for the difference between \MethodName and distance-based in the SRC of the Dunn Index and Handl-Knowles-Kell where $p<0.05$, and the difference between our method and CaD in the SRC of Bezdek-Pal where $p<0.05$. }

As can be seen by the experimental results presented in Tables \ref{tab:SRC_values_individual} and \ref{tab:MRR_values_individual} regarding the SRC and MRR results,  \MethodName  outperforms the other methods by a significant margin for all measures.  Based on these results,  we can infer that our proposed graphical embedded characterization is better at 
(1) identifying the rankings overall (measured by the SRC), as well as (2) recommending high-performing algorithms, where the lowest position of the first candidate is among the top-2 or top-3 algorithms (minimum MRR of $\sim$ {0.4}).

\subsubsection{Discussion: individual model results}

We compared the performance of \MethodName to that of the state-of-the-art methods (distance-based and CaD) and reached the following conclusions:

\textbf{\MethodName is a more robust approach.}
We analyzed six cases in which the improvement in the SRC measure of \MethodName and the CaD and distance-based methods was the greatest, in an attempt to identify the reasons for \MethodName's superior performance. 
The results of our analysis of the CaD and distance-based methods are summarized in Tables \ref{tab:differneces_individual_cad} and \ref{tab:differneces_individual_distance}, respectively. 
% It is clear that our approach is better suited for datasets with diverse properties (e.g., datasets that combine different types of features, have low/high-dimensionality (features and instances), and contain missing values).️
It is clear that our approach is better suited for datasets with diverse properties (e.g., datasets that combine different types of features, have low/high dimensionality (features and instances)).
These results demonstrate that \MethodName also addresses "challenging" and diverse datasets and is thus, more generic and robust. Furthermore, we argue that the results validate the core idea of our method: modeling the latent interactions among the instances of a dataset
provides a better representation for algorithm selection based on an individual measure than the predefined meta-features used by the CaD and distance-based methods. 

 \begin{table}[h!]
 \normalsize
    \centering
    \begin{adjustbox}{max width=\textwidth}
    \begin{tabular}{ l| l| l| l}
    \hline
        Index & Dataset Name & Improvement in SRC& Suggested Reason  \\
        \hline
         Bezdek-Pal & fri\_c3\_500\_50  & 0.670&Large number of features  \\
           Milligan-Cooper& mfeat-karhunen & 0.669& Large number of features\\
        Milligan-Cooper& seeds &   0.636 & Small number of instances \\
%           Xie-Beni& TKDSalesRegion4
%  &0.605 & Small number of instances\\
%          Dunn Index & fri\_c3\_500\_50 & 0.599  & Large number of features\\
         \hline
    \end{tabular}
    \end{adjustbox}
   \caption{ Individual model results - the top-3 datasets with the greatest improvement in the SRC measure between \MethodName and the CaD method. Positive values represent better performance by \MethodName. }
    \label{tab:differneces_individual_cad}
\end{table}

 \begin{table}[h!]
\Huge
    \centering
    \begin{adjustbox}{max width=\textwidth}
    \begin{tabular}{ l | l |l| l}
    \hline
        Index & Dataset Name & Improvement in SRC& Suggested Reason  \\
        \hline
        Milligan-Cooper &seeds & 0.639 &  Small number of instances \\
         Milligan-Cooper&TKDSalesRegion4  &0.629& \vtop{\hbox{\strut Combination of nominal and numeric features }\hbox{\strut + small number of instances}}\\
          Silhouette score&ESL  & 0.598 & Small number of features \\
        %   SD-Scat& autoUniv-au7-500 & 0.798& Combination of nominal and numeric features\\
        % Calinski-Harabasz& Echocardiogram &  0.755&  Missing values + combination of nominal and numeric features\\
        %  Dunn Index & fri\_c3\_500\_50 & 0.610 & Large number of features\\
         \hline
    \end{tabular}
    \end{adjustbox}
   \caption{Individual model results - the top-3 datasets with the greatest improvement in the SRC measure between \MethodName and the distance-based method. Positive values represent better performance by \MethodName.}
    \label{tab:differneces_individual_distance}
\end{table}
\textbf{\MethodName is consistently better at recommending the top-performing algorithms.}  
Similar to our analysis of the average ranking model, the results presented in Table \ref{tab:MRR@1} demonstrate that \MethodName is more consistent at recommending algorithms that are in first place on the ranked lists. Moreover, while both state-of-the-art methods were able to produce a correct recommendation for more than half of the datasets ($>$104) on two indices, our method was able to do so on {seven} of the 10 measures evaluated. 

\begin{table}[h]
    \centering
    \begin{tabular}{ l c c c}
    \hline
        Index & Distance-Based & CaD & \MethodName  \\
        \hline
         Bezdek-Pal & 109 & 105 &\textbf{118} \\
         Dunn Index & 29 & 39 &\textbf{111}\\
         Calinski-Harabasz& 23 &24  &\textbf{59}\\
         Silhouette score & 41 & 40 & \textbf{76}\\
         Milligan-Cooper & 20 &24&\textbf{43}\\
         Davies-Bouldin&  104 &106& {\textbf{141}}\\
         Handl-Knowles-Kell&77  &81& {\textbf{132}}\\
         Hubert-Levin& 22 &40 & \textbf{112}\\
         SD-Scat&  {57}& {64}& \textbf{107}\\
         Xie-Beni& {60}& {58} & \textbf{161}\\
         \hline
    \end{tabular}
   \caption{ Individual model results - the number of datasets for each approach for which the optimal algorithm was recommended. The best results are  highlighted.}
    \label{tab:MRR@1}
\end{table}

\textbf{\MethodName is consistently better than the Cad and distance-based methods.} 
In the previous point, we  evaluated the performance of \MethodName and both state-of-the-art methods based on their top recommendation (i.e., a single algorithm). Now, we examine the overall recommendation quality by evaluating the performance of the best top-K recommended algorithms. To do so, we analyzed the MRR@K (where K is the number of recommended algorithms considered) of \MethodName and the other methods on two internal measures (the {Calinski-Harabasz index} and the Milligan-Cooper index) on which our method obtained the worst results, as seen in Table \ref{tab:MRR@1}.

\begin{figure}[ht!]

\begin{subfigure}{0.5\textwidth}
\includegraphics[width=0.9\linewidth, height=5cm]{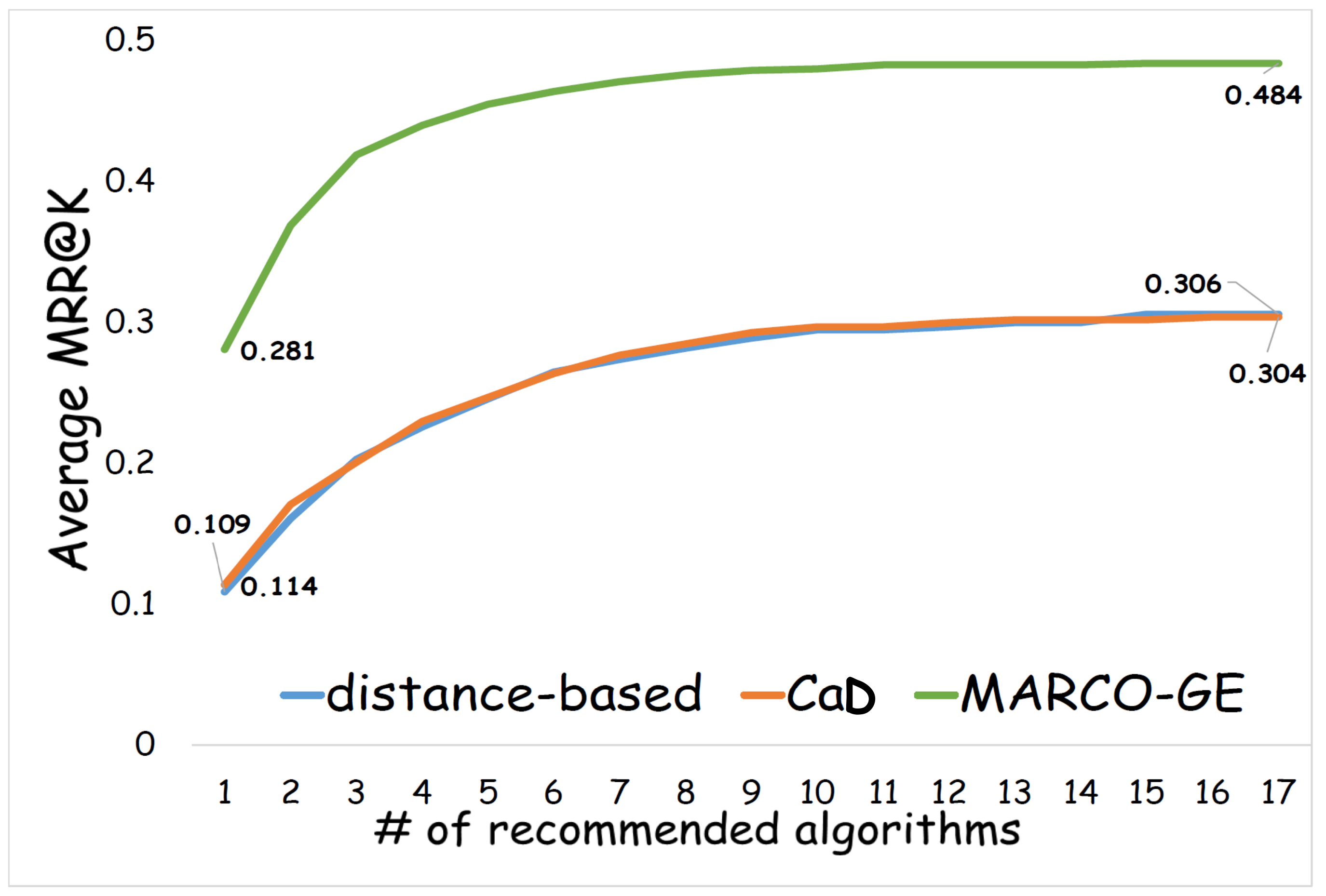} 
\caption{Average MRR@K for Calinski-Harabasz index.}
\label{fig:subim1}
\end{subfigure}
\begin{subfigure}{0.5\textwidth}
\includegraphics[width=\linewidth, height=5cm]{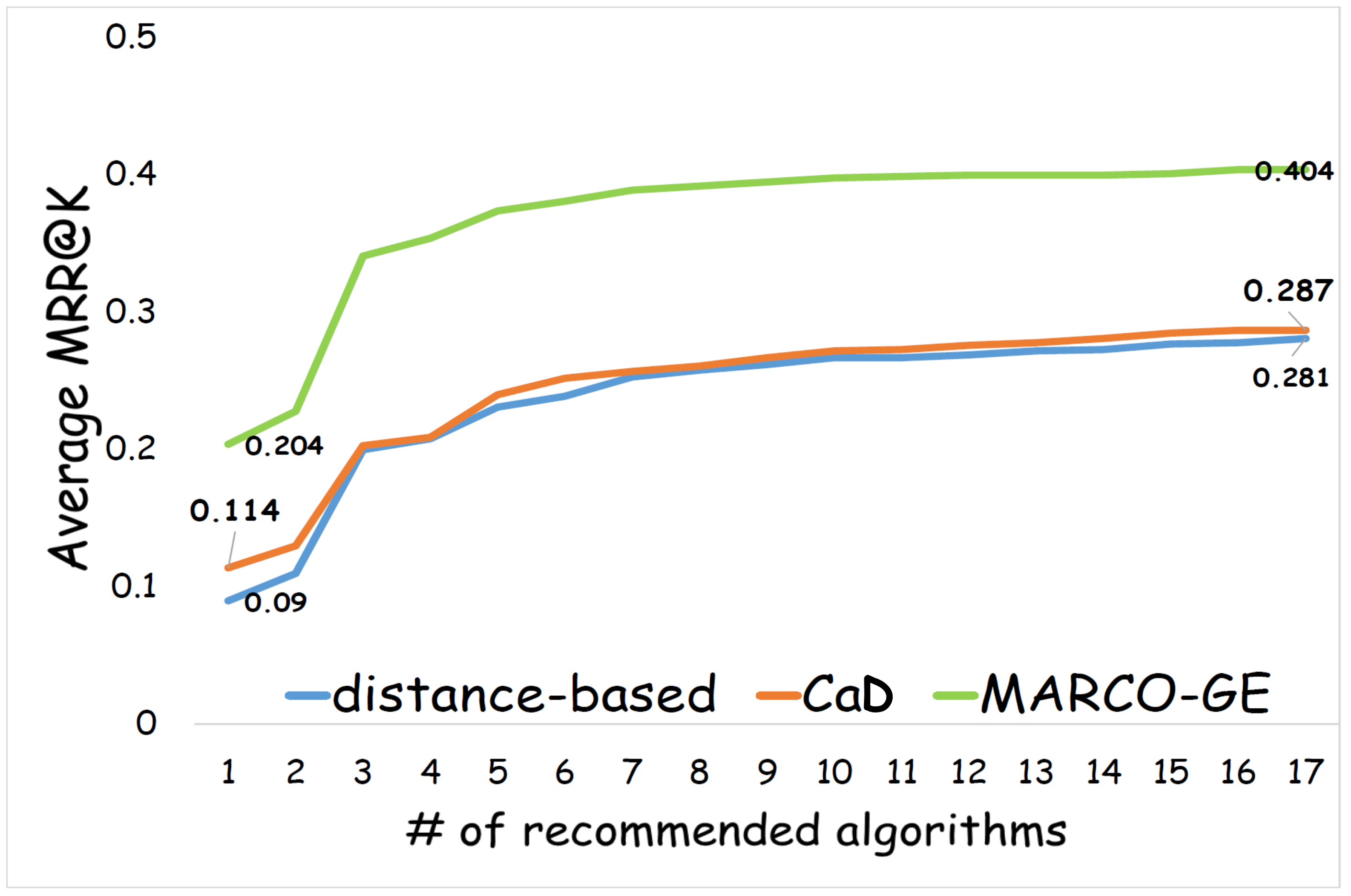}
\caption{Average MRR@K for Milligan-Cooper index.}
\label{fig:subim2}
\end{subfigure}

\caption{Individual model results - the average MRR@K over 210 datasets obtained by distance-based, CaD, and \MethodName vs the number of recommended algorithms.}
\label{fig:image2}
\end{figure}

The results, illustrated in Figure \ref{fig:image2}, show that \MethodName is consistently better than the CaD and distance-based methods. It can be seen that while the MRR@K value increases with the number of recommended algorithms for all methods,  \MethodName outperforms the other methods on the MRR@K metric by a wide and significant margin.  
These results indicate that our method is not only more consistent in its performance but that it can also avoid assigning unsuitable algorithms to a given dataset.

We can conclude that the ranked lists produced by \MethodName are effective, as they consist of multiple algorithms that achieve high performance for various datasets. 

To summarize the last two conclusions, we believe that the significant improvements in the MRR measure obtained by \MethodName derive from the supervised GCNN model used. Since the MRR measure obtains the maximum value when the best algorithm is ranked in the first position, and the GCNN model aims to detect the top-1 algorithm, the improvement in the MRR measure is more prominent than the SRC measure ({39.3\%} vs {6.4\%}, on average).

% which combines graph structure and graph label into an enriched embedding representation.
% MRR searches for the top-1 algorithm and  the supervised GCNN model.

%% file: Experiments/Popularity_results.tex
\subsection{Popularity-based baseline}
The popularity-based baseline proposed in this study is a ranking model based on the frequency of the algorithms as the best performers. 
\begin{table}[h!]
    \centering
    \begin{tabular}{l| c c| c c | c c | c c}
    \hline
   Index&  \multicolumn{2}{|c|}{Popularity-Based}& \multicolumn{2}{|c|}{Distance-Based} & \multicolumn{2}{|c|}{CaD} & \multicolumn{2}{c}{\MethodName}\\
    \hline
         &SRC& MRR& SRC &MRR & SRC &MRR&SRC&MRR  \\
      
         Bezdek-Pal& 0.380&{0.694}&0.492&0.628&0.500&0.618&{0.532}&0.679\\
         Dunn Index&0.345 &0.432&0.544&0.353&0.531&0.378&0.568&0.676\\
         Calinski-Harabasz&0.646&0.545&0.638&0.304&0.635&0.306&0.669&0.484\\
         Silhouette score& 0.405&0.406&0.487&0.368&0.487&0.370&0.527&0.529\\
         Milligan-Cooper& 0.327&0.390&0.428&0.281&0.410&0.287&0.456&0.404 \\
         Davies-Bouldin &{0.434}&{0.679}&{0.598}&{0.651}&{0.602}&{0.671}&{0.620}&{0.761} \\
         Handl-Knowles-Kell&{0.385}&{0.552}&{0.602}&{0.517}&{0.586}&{0.524}& {0.624}&{0.718} \\
         Hubert-Levin& {0.429}&{0.424}&{0.557}&{0.320}&{0.559}&{0.373}&{0.593}&{0.676}\\
         SD-Scat& {0.347}&{0.496}&{0.488}&{0.440}&{0.484}&{0.470}&{0.524}&{0.618}\\
         Xie-Beni& {0.313}&{0.518}&{0.502}&{0.482}&{0.502}&{0.471} &{0.526}&{0.828}\\
         \hline
            Average ranking& {0.409}&{0.513}&{0.631}&{0.509}&{0.629}&{0.499}&{0.645}&{0.822} \\
         \hline
         \end{tabular}
    \caption{The average SRC and MRR of the evaluated methods over 210 datasets and 11 internal clustering measures. }
    \label{tab:popularity_results}
\end{table}
The results, presented in Table \ref{tab:popularity_results}, show that  
\MethodName outperforms all of the evaluated methods, including the popularity-based approach, {except for the MRR of the Bezdek-Pal and Calinski-Harabasz indices where the popularity-based approach is better than \MethodName}. as seen in the table, the popularity-based approach provides reasonable performance and even demonstrates better MRR results than the other state-of-the-art methods (distance-based and CaD) on all of the  indices. Although the popularity-based approach does not require much computational effort (except for the evaluation of clustering algorithms' solutions) it is better at identifying the optimal algorithm.

%% file: Experiments/PCA.tex
\subsection{Analysis of the PCA technique's importance}
In this section we analyze the effect of using the PCA algorithm on \MethodName's performance. We conducted an ablation study with 210 datasets, using the same settings described in subsection \ref{setup}. Table \ref{tab:withandwithout} presents the performance results of the \textit{average ranking} model, with and without applying PCA. 

\begin{table}[h!]
    \centering
    \begin{tabular}{c c c}
    \hline
        Measure & \MethodName without PCA&  \MethodName
        with PCA\\
        \hline
         SRC& 0.631& \textbf{0.645}\\
         MRR& 0.774& \textbf{0.822}\\
         \hline
    \end{tabular}
    \caption{Average ranking model results - the average SRC and MRR values of \MethodName using two configurations: with PCA and without PCA. The mean values are computed based on 210 datasets. }
    \label{tab:withandwithout}
\end{table}

We used the Wilcoxon signed-rank test to validate the statistical significance of the differences between the corresponding values of SRC and MRR using two configurations: with PCA and without  PCA. The null hypothesis that our method  performs the same with and without PCA, and the observed differences are merely random was rejected with a significance level of 1\%  for both the SRC and MRR measures. 

Based on these results, we can conclude that the PCA algorithm contributes to the predictive performance of the proposed method.  Moreover, the PCA algorithm enables \MethodName to focus on the most informative edges in the graphs and disregard the remaining edges. This also results in reduced computational effort during the embedding calculation (decreased by a factor of three).

%% file: Experiments/Parameter_sensitivity.tex
\subsection{Hyperparameter sensitivity}
In these experiments, we investigate the influence of the configuration of the GCNN model on \MethodName's performance, in two respects: the model depth (number of graph convolutional layers) and graph embedding size.
\begin{figure}[h!] 
  \begin{subfigure}[b]{0.5\linewidth}
    \centering
    \includegraphics[width=0.9\linewidth]{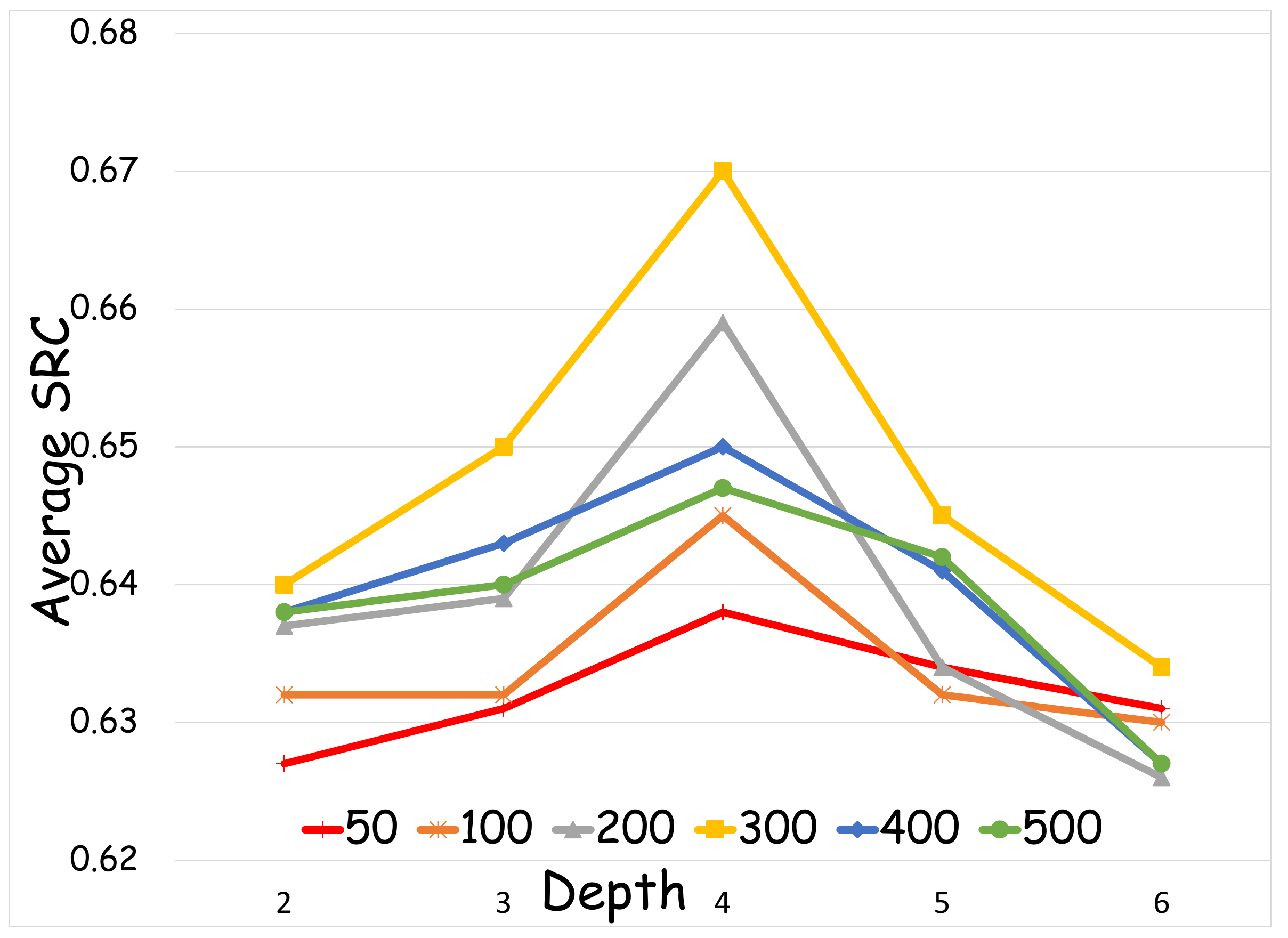} 
    \caption{Average ranking measure, SRC} 
    \label{fig7:a} 
    \vspace{4ex}
  \end{subfigure} %% 
  \begin{subfigure}[b]{0.5\linewidth}
    \centering
    \includegraphics[width=0.9\linewidth]{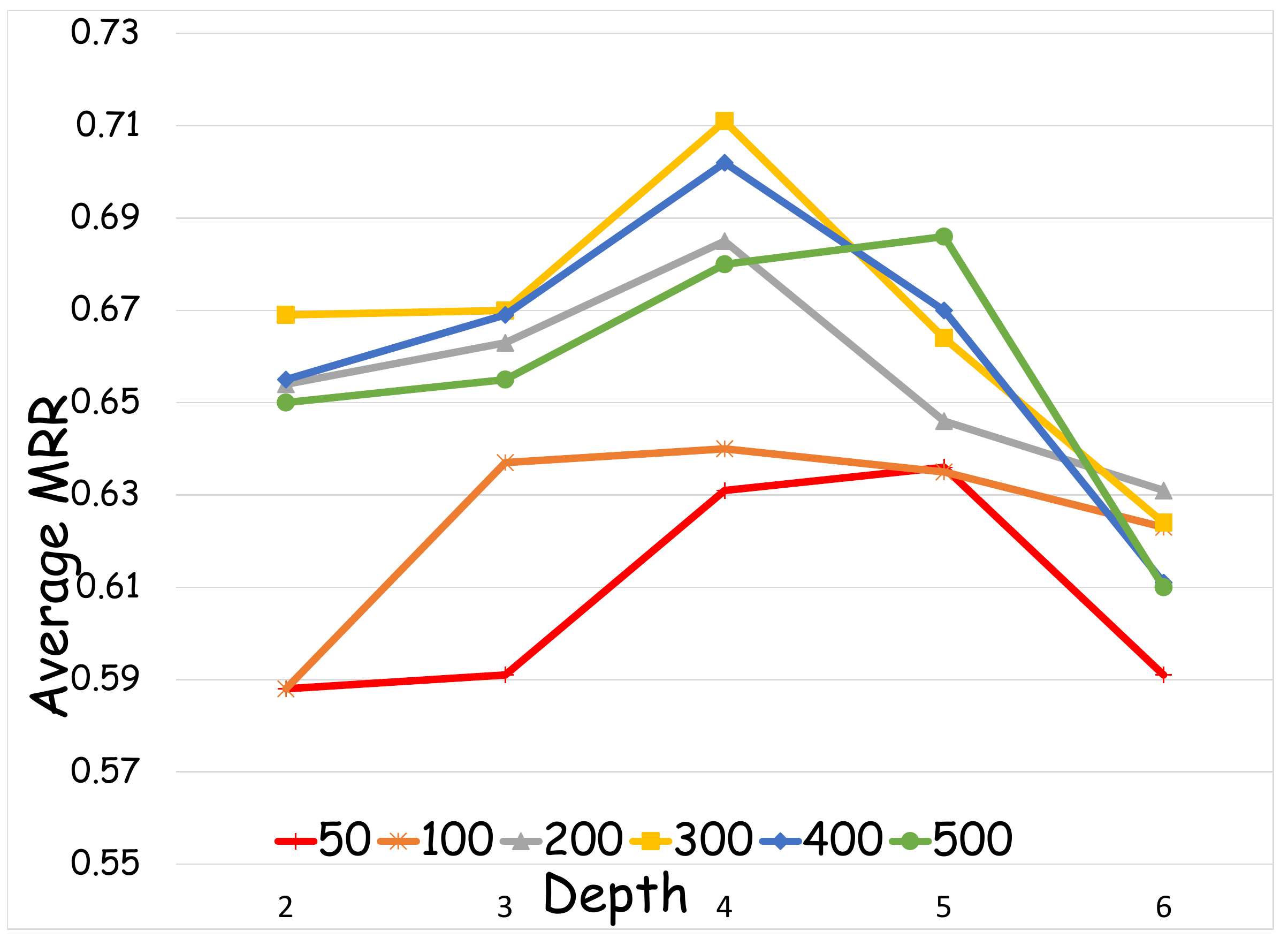} 
    \caption{Average ranking measure, MRR} 
    \label{fig7:b} 
    \vspace{4ex}
  \end{subfigure} 
  \begin{subfigure}[b]{0.5\linewidth}
    \centering
    \includegraphics[width=0.9\linewidth]{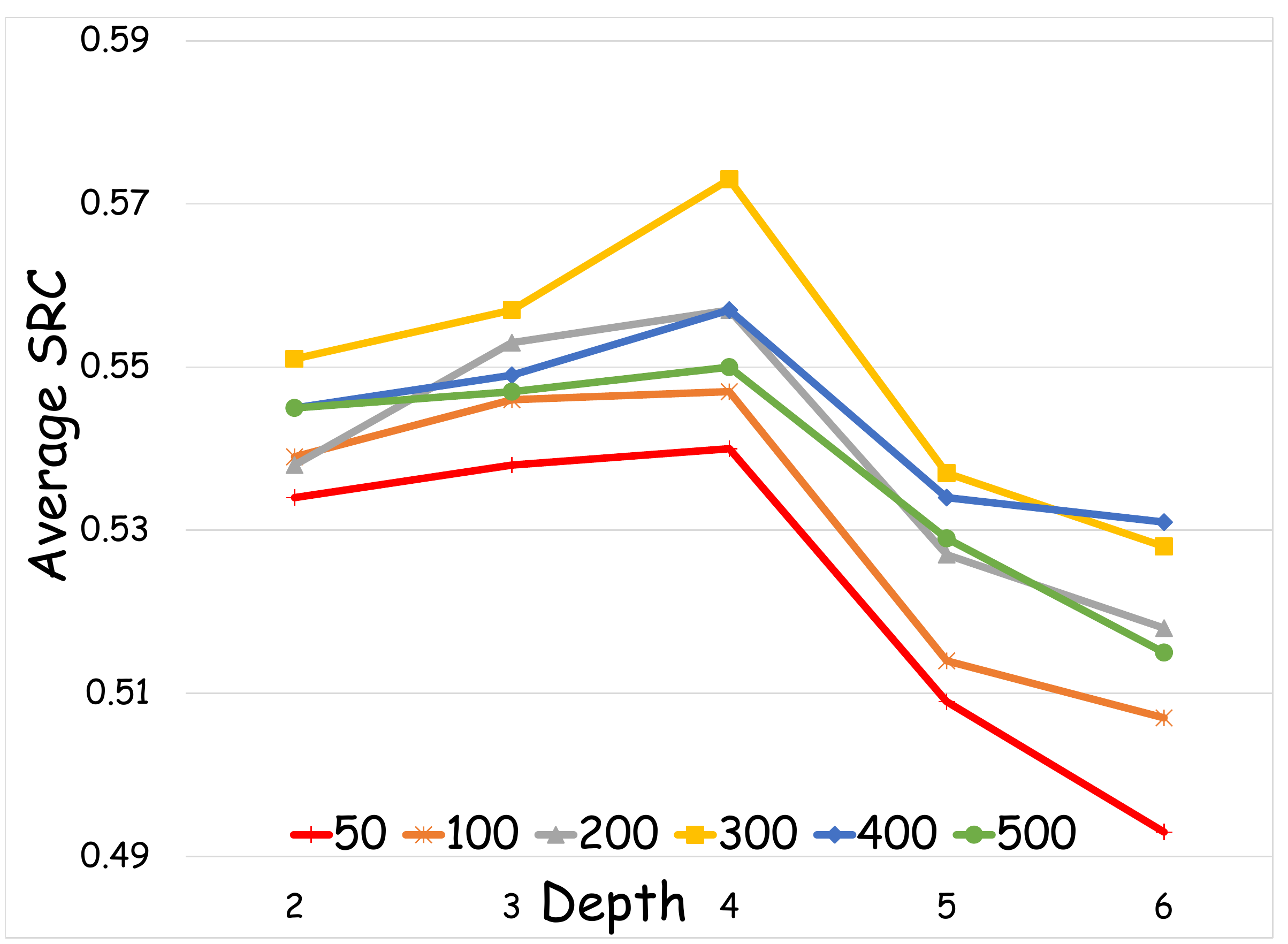} 
    \caption{Bezdek-Pal index, SRC} 
    \label{fig7:c} 
  \end{subfigure}%%
  \begin{subfigure}[b]{0.5\linewidth}
    \centering
    \includegraphics[width=0.9\linewidth]{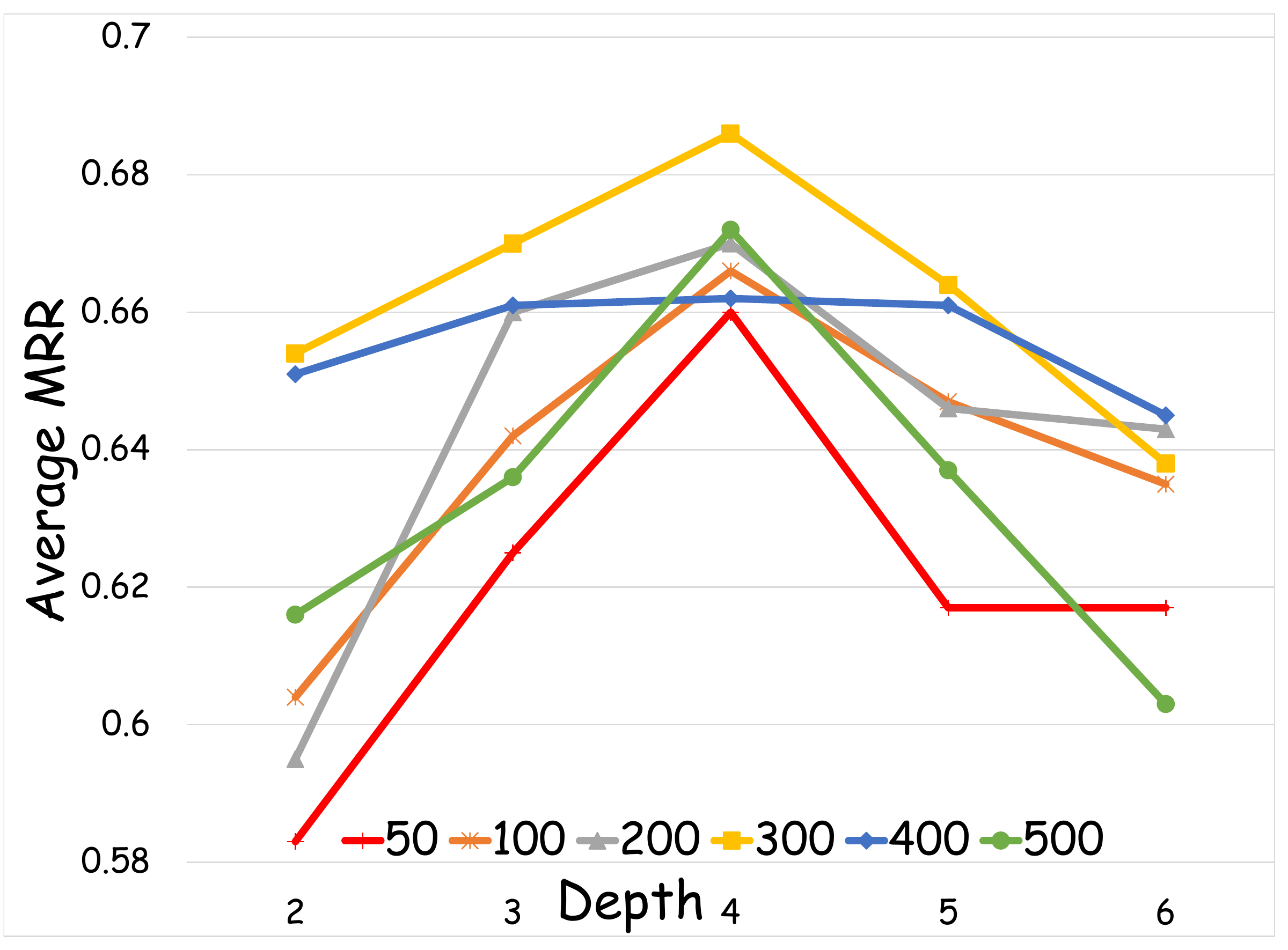} 
    \caption{Bezdek-Pal index, MRR} 
    \label{fig7:d} 
  \end{subfigure} 
  \caption{Influence of model depth (number of layers) and embedding size on \MethodName's performance. We present the SRC and MRR results  for both the average ranking measure and the Bezdek-Pal index. }
  \label{parameter_sen} 
\end{figure}
In order to evaluate how changes to the hyperparameters affect \MethodName's performance on the clustering algorithm selection task, we examined the average ranking measure and the Bezdek-Pal internal index. For these experiments, we randomly sampled 50 of the 210 datasets and set the number of epochs at 10. Except for the hyperparameter being tested, we used the values described in subsection \ref{setup} to set the parameters. We then varied the GCNN model depth from two to six layers and varied the size of the graph embedding using the following values \{50, 100, 200, 300, 400, 500\}. The effects of varying the embedding size and the model depth on the performance of \MethodName are shown in Figure \ref{parameter_sen}.

 From the figures, we can observe that as the embedding size and model depth increased from 50 to 300 and from two to four, respectively, the mean values of the SRC and MRR also increased, attaining their maximum with a four-layer model and an embedding size of 300.  
 In addition, in most cases, we can see that: (1) both of these parameters have a relatively high impact on \MethodName's  performance; (2)
 the performance of our method degrades if the model has more than four layers; (3) 
 our method obtains better results when utilizing high-dimensional vectors (300, 400, 500) than using low-dimensional vectors (50, 100); (4) using an embedding size of 300, the SRC measure achieved the best results for all depths, except for a depth of six, as seen in Figure \ref{fig7:c}; and (5) the best results for the MRR measure are obtained by using an embedding size of 300 and a depth of less than five.
 \input{Experiments/Computational_complexity}

%% file: Experiments/Computational_complexity.tex
{\subsection{Complexity analysis and computational cost}
In this section, we analyze the computational complexity of \MethodName and  its  run time. To obtain meta-features, our method applies the following three steps: \textit{clustering algorithm evaluation}, \textit{graph representation}, and \textit{meta-feature generation}. 
The computational complexity analysis focuses on the last two steps and disregards the first step (clustering algorithm evaluation), since it depends on the set of algorithms chosen and their implementations. (In subsection \ref{computational}, we report the computational time of all of the steps, including the \textit{clustering algorithm evaluation} step, based on the experiments performed and the set of algorithms presented in section \ref{clustering_algorithm}). 
 
% Finally, in the \textit{meta-model creation} step we train an \textbf{XGBoost} model. 

\subsubsection{Complexity analysis }
We investigate the complexity of the  meta-feature production process which includes five components; during the \textit{graph representation} step, three components are involved: \textbf{PCA computation}, \textbf{cosine similarity calculation}, and \textbf{graph construction}. Then, in the \textit{meta-feature generation} step, we apply the \textbf{DeepWalk} algorithm and train a \textbf{GCNN }model.
The computational complexity of each component is summarized in Table \ref{tab:complexity}.
Since in most cases we assume the number of features ($m$) $<$ number of instances ($n$), we can conclude that \MethodName's complexity is $O(n^2max(m+k))$.
} 

\begin{table}[h!]
    \centering
   { \begin{tabular}{l |l | l}
    \hline
       \textbf{\#}& \textbf{Component} & \textbf{Complexity}\\
        \hline
         1& PCA &  O($p^2q$) where $q$=min($n$,$m$), $p$=max($n$,$m$)\\
       2& Cosine similarity  & O($n^2m$) \\ 
       3& Graph construction & O($n^2)$ \\
       4& DeepWalk & O(n) \cite{pimentel2019efficient}\\ 
        5&GCNN model & O($kn^2$)  \cite{kipf2016semi} \\
         
      \hline  
    \end{tabular}}
    \caption{{The computational complexity of  \MethodName's components ($n$ - number of instances in  dataset, $m$ - number of features in dataset, and $k$ - the number of datasets).}}
    \label{tab:complexity}
\end{table}

{
\subsubsection{Computational cost}\label{computational}
In order to assess the processing time for obtaining the meta-features, we distinguish between the \textit{training} and \textit{inference} phases. 
\subsubsubsection{Training phase}
During the training phase, three steps are performed: clustering algorithm evaluation, graph representation, and meta-feature generation. 
The first step involves: (a) running the clustering algorithms on a collection of datasets, and (b) evaluating their solutions using 10 internal measures. 
Table \ref{tab:run_time_algorithms} presents the minimum, average, and maximum run time of each algorithm over all 210 datasets. It can be seen that the average run time of a large majority of the algorithms can be disregarded; however, for algorithms EAC, PSC, and KHM, the average run time is much longer than the others, and in these cases it must be considered. Each algorithm produces a clustering solution that is evaluated by internal measures. Table \ref{tab:run_time_indices} summarizes the average computing time of each measure for each algorithm over all of the datasets. Again, one can see that there are indices that have a longer run time (i.e., Milligan-Cooper, Hubert-Levin), meriting consideration. 
\begin{table}[h!]
    \centering
    {\begin{tabular}{l l l l }
    \hline
         \textbf{Algorithm}& \textbf{Minimum run time}&\textbf{Average run time} & \textbf{Maximun run time}   \\
         \hline
         EAC& 2776.5&105978.5  & 2532184.7\\
         PSC& 1027.3 & 184295.8& 3910181.1\\
         MST&6.8 & 40.1 & 2047.8\\
         SL& 3.42 & 20.9 & 694.3\\
         AL& 2.9& 25.9 & 539.2\\
         CL& 3.5&25.0 & 520.6\\
         WL& 3.5& 26.0& 516.5\\
         KM& 16.2&111.1 & 1213.4\\
         KHM& 414.4& 66623.4 & 1254226.9\\
         KKM& 11.0& 117.9 & 2745.3 \\
         MBK & 12.1& 50.1& 147.7 \\
         FC &5.9& 35.8 & 1000.0 \\
         DBSCAN &4.8& 60.7&  2779.8\\
         MS& 34.0& 744.6& 9123.2\\
         GMF& 8.3& 63.0& 465.5\\
         GMT & 6.3& 48.0& 754.6\\
         GMD &6.1& 32.8 & 278.0\\
         \hline
    \end{tabular}}
    \caption{{The minimum, average, and maximum run time (in milliseconds) of the evaluated clustering algorithms over 210 datasets.}}
    \label{tab:run_time_algorithms}

\end{table}

\begin{table}[h!]
\Huge
    \centering
    \begin{adjustbox}{max width=\textwidth}
    {\begin{tabular}{l l l l l l l l l l l l l l l l l l  }
    \hline
   Index&  EAC & PSC & MST & SL & AL & CL & WL & KM & KHM & KKM &  MBK & FC & DBSCAN & MS & GMF & GMT & GMD \\
    \hline
    
         Bezdek-Pal& 352.6&271.9&52.9&210.5&279.8&442.1&334.9&381.7&326.7&139.5&303.5&186.9& 160.0& 43730.0& 333.6& 312.3&  312.3\\
         Dunn Index& 418.3 &320.7&77.0&304.6&359.3&493.7&393.1&445.3 & 373.9&194.2& 354.9&240.0& 237.0& 43740.5& 392.0& 377.7& 383.6\\
         Calinski-Harabasz&2.7&2.8&0.8&2.6&2.8&2.9&2.7&2.8&2.7& 2.3& 2.7&2.5&2.3& 0.9& 2.8& 2.7& 2.7\\
         Silhouette score& 15.6&16.18&6.5&15.8&15.6&14.9&15.4&15.4& 15.9& 14.8&  15.2& 15.1& 15.4& 1.4& 14.8& 15.0 & 14.5\\
         Milligan-Cooper& 12655.6&12631.0&5191.8&12728.2&12628.1&12528.2&12440.7&12664.5& 12713.6& 12220.8& 12487.1& 12462.5& 11090.7& 1011.3 & 12140.2& 12137.5& 12096.3\\
         Davies-Bouldin &5.5&6.1&1.6&5.1&5.&6.4&5.6&6.1& 5.6& 3.7& 5.6& 4.4 & 4.9& 6.1 & 5.8&5.7&5.5\\
        Handl-Knowles-Kell&239.1&241.8&93.5&238.9&226.0&230.7& 237.9&241.8& 232.2&  242.0& 245.9& 233.95& 205.2&25.6& 229.2& 224.8& 223.6\\
         Hubert-Levin& 53337.8&60115.8&23998.5&42529.4&46843.4&42105.2&63607.8&62009.4& 58131.1& 62929.4& 56236.5& 65132.4& 48968.1&946.6& 54836.5& 54953.3& 51670.1\\
         SD-Scat& 2.3&2.4&0.6&2.2&2.5&2.9&2.3&2.6& 2.2& 1.4& 2.4& 1.7&1.6&8.3&2.4&2.4& 2.3\\
         Xie-Beni& 68.1&75.6&25.1&62.1&67.2&114.7 &79.6&84.5& 65.0& 59.5& 75.6&56.0& 40.3& 2182.1& 71.9 & 72.5& 69.7\\
         
         \hline
         \end{tabular}}
         \end{adjustbox}
    \caption{{The average run time (in milliseconds) for computing the internal measures for each algorithm. Each value represents the mean result over 210 datasets.}}
    \label{tab:run_time_indices}
\end{table}

\begin{figure}[h!]
    \centering
    \begin{adjustbox}{max width=1\textwidth}
    \includegraphics{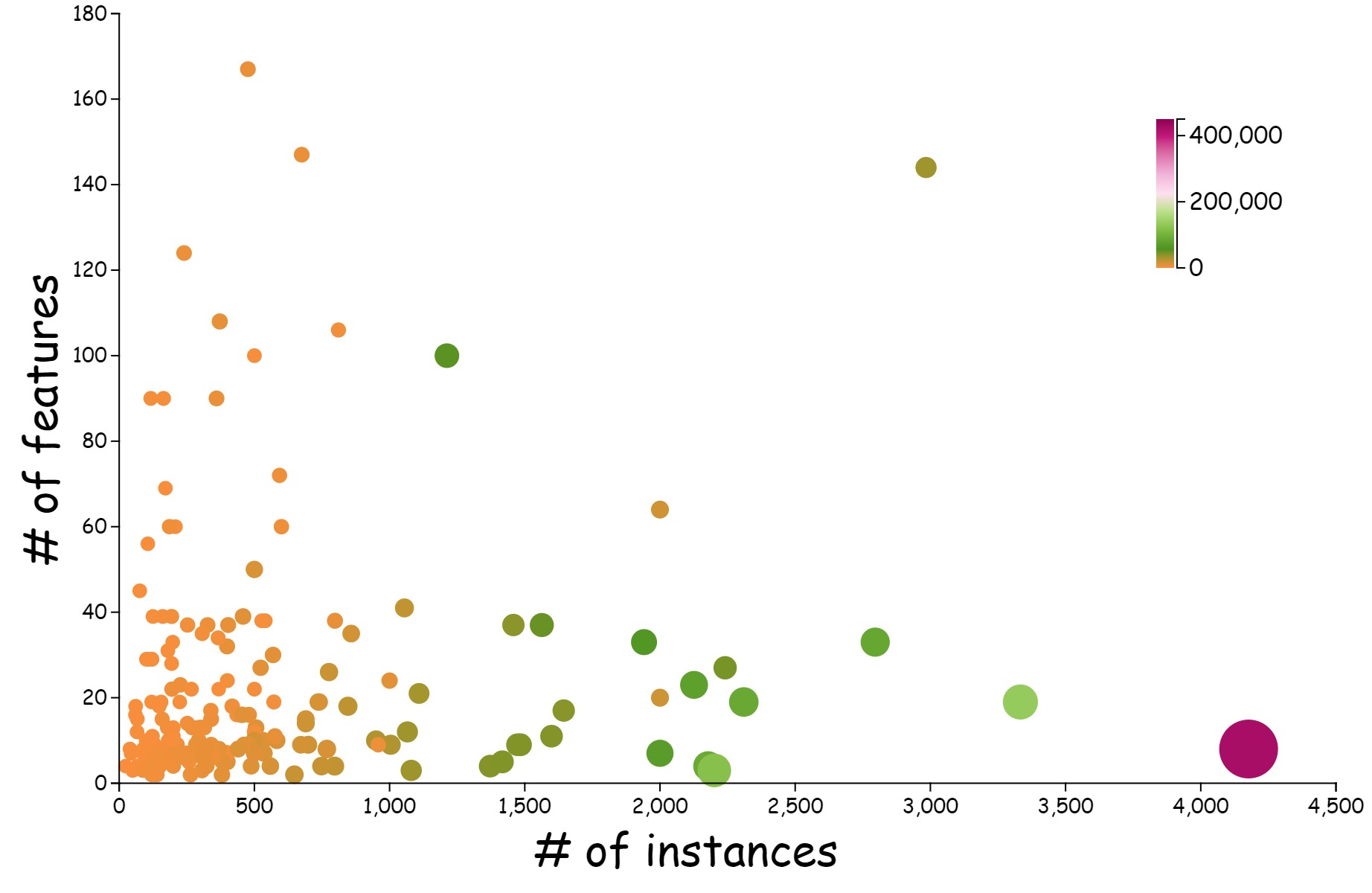}
    \end{adjustbox}
    \caption{{The total execution time (in milliseconds) of components 1-4 for all 210 datasets in relation to the datasets' dimensions. The size of each bubble represents the total processing time.  }}
    \label{fig:run_time}
\end{figure}

The next steps to be analyzed are graph representation and meta-feature generation. Together, these steps  include  components 1-5 from Table \ref{tab:complexity}.  We merged the computation times of components 1-4 (i.e., PCA,  cosine similarity, graph construction, and DeepWalk) and used a bubble chart (Figure \ref{fig:run_time}) to present the effect of the datasets' dimensions on the processing time of these components. The size of each bubble in Figure \ref{fig:run_time}  represents the total execution time of components 1-4. Although it can be observed that the run time of these components is influenced by the number of instances, the processing time of the dataset with the largest number of instances is about five minutes. 

Afterward, a GCNN model is trained on all of the datasets. The average time to train the GCNN model is around an hour (for all of the datasets together).

% It is important to note that the training phase, and particularly the steps of algorithm evaluation and training the GCNN model, only need to be done once during the training phase and not at all in the inference phase, when new datasets arrive.  

It is important to note that when new datasets arrive, the training phase steps of algorithm evaluation and GCNN model training only need to be done once during the training phase and not at all in the inference phase.

\subsubsubsection{The inference phase}
In order to produce a recommendation for the best-performing algorithm for an unseen dataset,  the following steps are required: (a) PCA computation, (b) cosine similarity calculation, (c) graph construction, (d) DeepWalk calculation, and (e) utilizing the trained GCNN model to obtain the meta-features for the dataset at hand. The execution time (presented in Figure \ref{fig:run_time}) of steps a-d, which are actually components 1-4 from Table \ref{tab:complexity}, is identical to the computational time of these components during the training phase. Step (e) takes about 10 milliseconds per dataset. 
Once steps a-e are completed, these features are fed into the meta-learner model (i.e.,  XGBoost),  which in turn generates a recommendation for the best-performing algorithm and its corresponding best configuration (according to the hyperparameter tuning process described in subsection \ref{hyperparameter_tuning}). One can then test this particular algorithm  on the dataset in question. 

% While the training phase includes evaluation of the clustering algorithms and training of the GCNN model, the inference phase uses the obtained model for generating the meta-features for the dataset at hand. Consequently, the processing time of a particular dataset for the other components: PCA, cosine similarity, graph construction, and DeepWalk is equivalent for both the training and the inference. In order to present how the dataset's dimensions effects the 

% \subsubsubsection{Algorithms run time}
% In this subsection the run time of the 17 clustering algorithms presented above is provided.  Table XX summarizes the average running time of each algorithm over all the datasets (i.e., 210). 

}

%% file: Conclusions.tex
\section{Conclusions and future work}
 In this study, we introduced \MethodName, a meta-learning method aimed at recommending the best clustering algorithms for a given dataset and evaluation measure. By modeling the interactions of the dataset's instances as a graph and extracting an embedding representation that serves the same function as  meta-features, we were able to develop a meta-learning model capable of effectively recommending top-performing algorithms for previously unseen datasets. Our proposed approach outperforms leading existing solutions such as the CaD and distance-based methods,  while also proving itself highly effective with "challenging" datasets (e.g., those with high-dimensionality) and on the task of identifying the optimal algorithm in the first rank position.
 
 For future work, new meta-features representing additional elements of the datasets can be extracted from the graphical representation of the dataset.
 In addition, the robustness of the meta-model can be improved by including more datasets. 
 
\section{Acknowledgments}
This research was partially supported by the Israel's Council for Higher Education (CHE) via the Data
Science Research Center at Ben-Gurion University of the Negev, Israel.